\documentclass[runningheads]{llncs}

\usepackage{eccv}

\usepackage{eccvabbrv}

\usepackage{graphicx}
\usepackage{booktabs}
\usepackage{adjustbox}
\usepackage{siunitx}   
\usepackage{array}      
\usepackage{multirow}
\usepackage[accsupp]{axessibility}

\usepackage{hyperref}
\begin{document}

\title{PWM-ArtGen: Part World Model for Articulated Object Generation} 

\titlerunning{PWM-ArtGen: Part World Model for Articulated Object Generation}

\author{
Wentao Zheng
\and
Ancong Wu\thanks{Corresponding author.}
}

\authorrunning{W.~Zheng and A.~Wu}

\institute{School of Computer Science and Engineering, Sun Yat-sen University, China\\
\email{zhengwt28@mail2.sysu.edu.cn, wuanc@mail.sysu.edu.cn}
}

\maketitle

\begin{abstract}
The key challenge in articulated 3D object generation from a single image is accurately predicting the underlying kinematic structure.
Existing methods either infer kinematic parameters directly from a static image that lacks dynamic part-level kinematic relationships, or estimate parameters from visual dynamics generated from a single image, which is prone to accumulated errors of two steps. Moreover, the limited scale and diversity of existing annotated datasets further hinder generalization to complex, real-world objects. To overcome these limitations, we propose to learn the joint distribution of visual dynamics and kinematic parameters. Recognizing that articulated objects can be formulated as dynamic systems, we propose a unified Part World Model called \textbf{PWM-ArtGen}. To leverage unannotated data, this model couples action diffusion and image diffusion with independent diffusion timesteps, which enables visual branch co-training.
We further curate a photorealistic dataset of 19.7k part-level image pairs without kinematic annotations, to support co-training. Experiments demonstrate that PWM-ArtGen substantially outperforms existing baselines in the resting state and exhibits strong zero-shot generalization to out-of-distribution objects. We will release our code at \url{https://github.com/Wentap123/PWM-ArtGen}.
  \keywords{articulated object \and visual dynamics \and kinematic parameters}
\end{abstract}

\section{Introduction}
\label{sec:intro}

Articulated household objects are ubiquitous in daily environments and serve as fundamental building blocks for interactive virtual worlds, robotics, and embodied AI~\cite{shen2021igibson, wu2026dexgrasp}. Automatically creating high-quality articulated assets at scale remains challenging: expert authoring is costly, and data capture in real settings is often constrained. As a result, increasing attention has been devoted to developing automatic methods for articulated object modeling. 

Existing methods fall into two camps: one line of work focuses on optimization-based reconstruction, assumes multi-image or video inputs~\cite{Liu_ICCV_2023_PARIS,liu2025videoartgs,liu2025artgs,peng2025itaco} are available or synthesizes multi-images or video~\cite{li2024dragapart,li2025puppet,lu2025dreamart} from a single input image. Such approaches inject motion cues and connectivity information to recover articulated behavior, but they often rely on heavy optimization. They also require careful cross-view or cross-state alignment, or explicit capture of the object in motion, which is not always feasible in practice and limits throughput and scalability to diverse real-world objects.
Other approaches~\cite{Liu_CVPR_2024_CAGE,Liu_ICLR_2025_SINGAPO} regress kinematic parameters directly using trainable feedforward networks trained on existing datasets. While they can produce plausible geometry and kinematics from a resting-state image, they exhibit clear performance degradation when applied to structurally complex or visually ambiguous objects. Since a single static image provides no explicit articulation cues, the absence of dynamic information fundamentally limits identifiability.

Beyond methodological constraints, progress is bottlenecked by the scarcity of high-quality training data. Synthetic datasets like PartNet-Mobility (PM)~\cite{Mo_CVPR_2019_PartNet,Xiang_CVPR_2020_SAPIEN} lack photorealism, often leading to poor real-world generalization. Conversely, realistic datasets like Articulated Containers
Dataset (ACD)~\cite{iliash2026s2o,Collins_CVPR_2022_ABO} are too small for large-scale training and are better suited for evaluation. Since acquiring kinematic annotations for real-world objects is prohibitively expensive, exploring methods that can leverage unannotated visual data for training is of great significance. For such unannotated data, we can harness off-the-shelf image and video foundation models to generate or enhance visual priors, easily scaling up diverse, high-quality datasets without relying on costly manual labels.

Motivated by the challenges of single-image articulated generation on real data, we argue that prior work~\cite{Liu_CVPR_2024_CAGE,Liu_ICLR_2025_SINGAPO} largely neglects part-level appearance dynamics and instead focuses on kinematic parameters. Our key insight is that visual dynamics and kinematics are inherently inter-dependent: observed motions constrain feasible joint types and axes, while kinematic parameters explain motion patterns.
We therefore introduce \textbf{PWM-ArtGen}, a generative model that learns the joint distribution of part-level visual dynamics and kinematic parameters. 
By decoupling diffusion timesteps, our approach facilitates visual-branch co-training on unannotated data, thereby capturing complex visual dynamics without the need for manual labels.

Also, we introduce a Visual Dynamics Regularizer (VDR) that explicitly promotes faithful part-level dynamics, thereby limiting dependence on static cues present in a single image.
To support co-training, we curate PartNet-Mobility-Reality (PM-R), a photorealistic dataset of 19.7k paired images. Concretely, we apply a frozen image-to-image editing model to PartNet-Mobility renders to enhance materials, lighting, and backgrounds.
In summary, our main contributions are as follows:
\begin{itemize}
    \item \textbf{PWM-ArtGen:} We propose a single-image part world model that jointly learns visual dynamics and kinematics, enabling controllable 3D articulated generation and seamless co-training on unannotated data.
    \item \textbf{Visual Dynamics Regularizer (VDR):} We introduce a regularization mechanism that enforces consistency between visual motion cues and articulated behaviors, significantly improving kinematic reasoning.
    \item \textbf{PM-R Dataset:} We construct a photorealistic dataset of 19.7k image pairs with part-level dynamics, which unlocks scalable co-training and effectively bridges the synthetic-to-real gap.
\end{itemize}

\section{Related Work}
\label{sec:relatedwork}

\subsection{Articulated object creation}
\noindent \textit{\textbf{Reconstruction-based Methods.}}
To advance robotic simulation and digital twins, the rapid generation of articulated object assets from images has garnered significant interest. Forming the foundation of this domain, initial works~\cite{hu2017learning,wang2019shape2motion,yan2019rpm,geng2023gapartnet,Mo_CVPR_2019_PartNet,Xiang_CVPR_2020_SAPIEN,iliash2026s2o,Collins_CVPR_2022_ABO} largely depended on manual annotation to construct extensive datasets. One prominent line of work recovers articulated behavior through per-object optimization. Early approaches in this domain learn deformable signed distance fields~\cite{Mu_ICCV_2021_ASDF} or reconstruct parts from point clouds captured across multiple object states~\cite{Jiang_CVPR_2022_Ditto,Liu_CVPR_2023_Rearticulable,Tseng_ICRA_2022_CLANeRF}. To inject essential motion cues and connectivity information, subsequent methods rely heavily on multi-view imagery, stereo captures~\cite{Heppert_CVPR_2023_Carto}, or video inputs~\cite{Liu_ICCV_2023_PARIS, liu2025videoartgs, liu2025artgs, peng2025itaco, Weng_CVPR_2024_NeuralImplicit, Song_CVPR_2024_Reacto}. Recent advancements even attempt to synthesize these multi-view or video inputs from a single image to guide the optimization process~\cite{li2024dragapart, li2025puppet, lu2025dreamart}. However, these approaches consistently suffer from low optimization speed, fixed-category assumptions, and a strict dependence on dense cross-view or cross-state alignment. This limits their practical throughput and scalability to diverse, in-the-wild objects.

\noindent \textit{\textbf{Generation-based Methods.}}
Another line of work~\cite{Liu_CVPR_2024_CAGE,Liu_ICLR_2025_SINGAPO,Chen_RSS_2024_URDFormer,le2025articulate} synthesizes articulated objects from a single image by assembling or adapting parts retrieved from pre-constructed asset collections. Although such retrieval-based paradigms benefit from realistic part priors, the absence of dynamic cues fundamentally limits their ability to infer complex real-world articulations. Recent work~\cite{wudipo} begins to incorporate motion-related observations, but it relies on dual-image input and therefore cannot be directly applied to the more challenging single-image setting.
Beyond retrieval-based generation, several methods~\cite{Lei_NeurIPS_2023_NAP,Su_CVPR_2025_Artformer} attempt to synthesize articulated geometry through implicit representations such as generated SDFs followed by surface reconstruction. However, the reconstruction process can introduce artifacts and often struggles to preserve fine-grained geometric details. Recent advances in 3D foundation models have further enabled generative articulated object modeling~\cite{chen2025freeart3d,chen2025artilatent}, yet existing solutions are still primarily demonstrated on relatively simple articulation patterns. 

Unlike prior single-image methods that focus solely on static geometry or direct parameter regression and thereby overlook dynamic information, we argue that visual dynamics provide crucial physical constraints for kinematic reasoning. PWM-ArtGen jointly models visual dynamics and kinematic parameters. This coupled formulation not only improves parameter estimation but also enables effective co-training on large-scale unannotated data, making our method scalable and generalizable to complex real-world objects.

\subsection{World Model}
Recent advancements in world models~\cite{brooks2024video, bruce2024genie, bardesrevisiting, agarwal2025cosmos} have demonstrated remarkable success in learning general physical dynamics from large-scale unannotated videos, which significantly enhances the reasoning and prediction of actions. Building on this, recent approaches~\cite{Zhu_RSS_2025_UWM, wuunleashing, guo2024prediction} have effectively harnessed video representations to facilitate the prediction of complex robotic trajectories and control actions. Despite these successes, the introduction of world models into the domain of articulated object generation remains largely unexplored, yet it presents a highly promising avenue for capturing intricate structural and kinematic priors. Furthermore, while existing frameworks,~\ie UWM~\cite{Zhu_RSS_2025_UWM}, are inherently designed for control policies and cannot be directly applied to this generation task, we adapt their core principles to seamlessly fit the unique requirements of articulated object generation.

\section{Part World Model}
\label{sec:method}
\subsection{Overview}
\label{sec:overview}
\begin{figure}[t] 
  \centering
  \includegraphics[width=\textwidth, trim=0cm 3cm 0cm 2cm, clip]{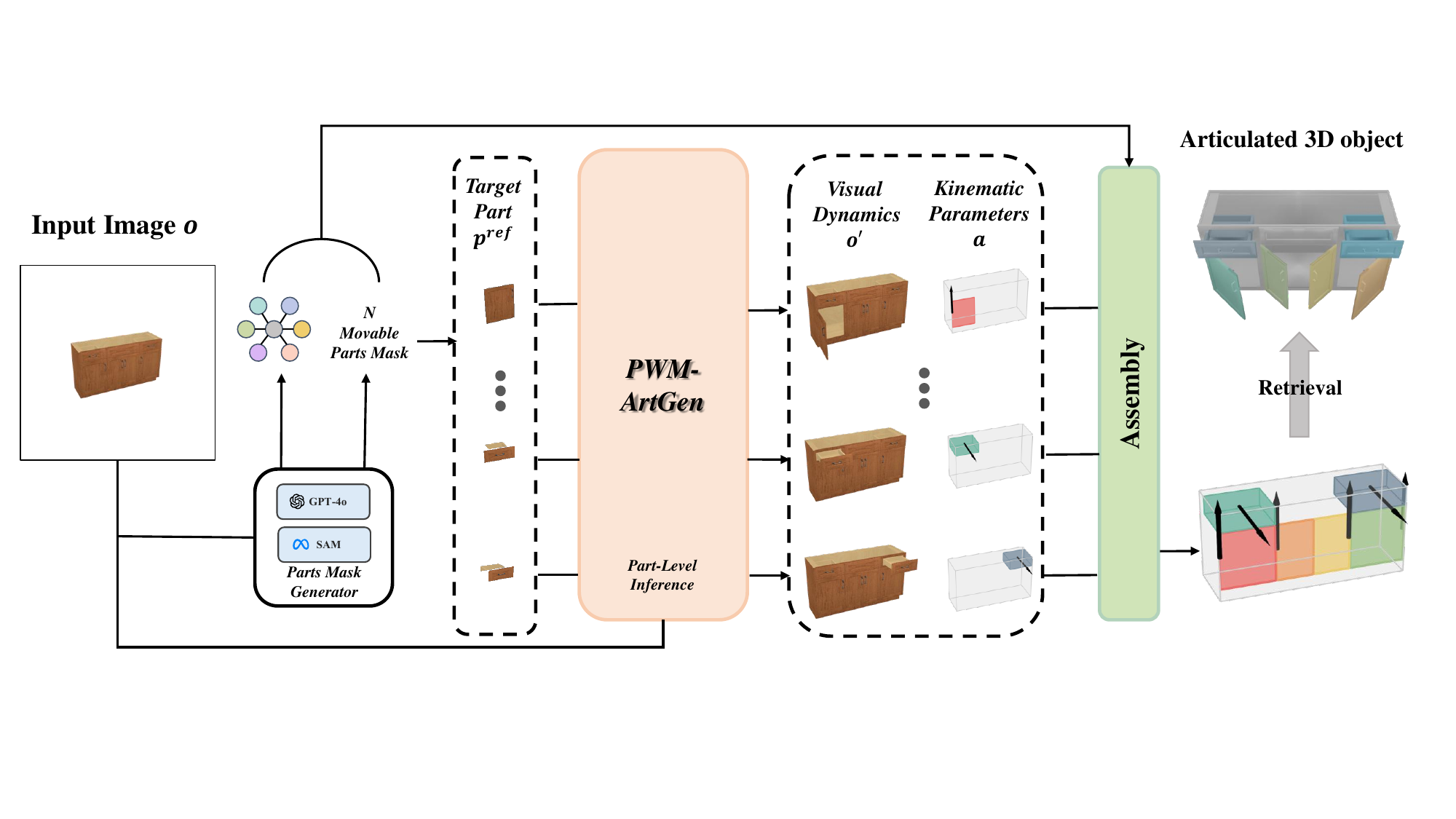} 
\caption{Overview of PWM-ArtGen, an end-to-end pipeline to convert a single image into an articulated 3D object: (1) \textbf{Parts mask generator}: from image $o$, a pretrained module based on SAM~\cite{kirillov2023segment} and GPT-4o~\cite{hurst2024gpt}, extracts $N$ part masks $\{m_i\}_{i=1}^N$ and an articulate graph for assembly, (2) \textbf{Part-level inference}: $p^{\text{ref}}_i$ is instantiated as the part mask $m_i$, sample $(o'_i, a_i)\sim p(o', a \mid o,\, p^{ref}_i)$, (3) \textbf{Assembly}: align and assemble $\{(o'_i, a_i)\}_{i=1}^N$ with the shared base bounding box $\mathbf{b}^{\mathit{base}}$ in the object, (4) \textbf{Retrieval}: following prior work~\cite{Liu_CVPR_2024_CAGE,Liu_ICLR_2025_SINGAPO} to regularize geometry and motion. Implementation details of the \textit{Part Mask Generator}, \textit{Assembly}, and \textit{Retrieval} are provided in the Supplementary Material.
}
  \label{fig:pipeline}
\end{figure}
We propose a diffusion-based Part World Model that learns the joint distribution of kinematic attributes and dynamic visual representations conditioned on the input image and target articulated part. Drawing on UWM~\cite{Zhu_RSS_2025_UWM}, we utilize independent diffusion timesteps to ingest action-free observations. This strategy allows the visual branch to extract kinematically relevant features from unannotated data, significantly enhancing kinematic estimation accuracy.
We adopt an articulation-centric parameterization and introduce a visual dynamics regularizer that explicitly promotes faithful part-level dynamics.

Formally, the model learns the conditional distribution $p(o', a \mid o, \;p^{ref})$ through a coupled noise prediction network $s_\theta$, where $o$ denotes the static observation, $o'$ the post-action dynamic observation, $a$ the kinematic attributes, and $p^{ref}$ the target articulated part serving as structural guidance. For each input image, the model generates coherent part-level dynamics and motion parameters that are spatially organized according to $p^{ref}$, enabling consistent reconstruction of articulated objects. Based on the model, we construct a pipeline that assembles parts into an object generation by retrieval, which is illustrated in~\cref{fig:pipeline}. 
\subsection{Data Representation}
\label{sec:REPRESENTATION}
We represent each articulated part with a dynamic visual, a set of kinematic attributes, and corresponding part-level structural guidance.

\noindent \textit{\textbf{Dynamic visual.}}
The post-action dynamic observation $o'$ is a $256 \times 256$ RGB image depicting the scene after executing the action on the reference part.

\noindent \textit{\textbf{Kinematic attributes.}}
Compared to prior work~\cite{Liu_CVPR_2024_CAGE,Liu_ICLR_2025_SINGAPO}, we expand the kinematic attributes with $a = \{t, r, \mathbf{l}, \mathbf{d}, \mathbf{b}, \mathbf{b}^{\mathit{base}}\}$, where $t$ denotes the joint type, $r$ the motion range, $\mathbf{l} \in \mathbb{R}^3$ the joint axis location, $\mathbf{d} \in \mathbb{R}^3$ the joint axis direction, $\mathbf{b} \in \mathbb{R}^6$ the part bounding box, and $\mathbf{b}^{\mathit{base}} \in \mathbb{R}^6$ the base bounding box. Considering dynamic visual behavior, the articulation types addressed in this work include revolute and prismatic joints. All articulated parts of the same object share the base bounding box $\mathbf{b}^{\mathit{base}}$, which provides a consistent reference for post-assembly in our part-level modeling framework. This design eliminates the need to predefine a maximum number of parts, as required by prior works. 

\noindent \textit{\textbf{Part-level structural guidance.}}
The target articulated part $p^{ref}$ is derived from a part mask and comprises both the cropped part image and its 2D geometric cues:
\begin{equation}
p^{\text{ref}} = \{o^{\text{part}},\, \text{2Dbbox}^{\text{part}},\, \text{2Dbbox}^{\text{base}}\}.
\label{eq:pref}
\end{equation}
${o}^{part}$ means the cropped part observation by the part mask, and ${2Dbbox}^{{part}}$, ${2Dbbox}^{{base}}$ encode the spatial extent of the region within the image using its center coordinates $(c_x, c_y)$, height $h$, and width $w$. We adapt DINOv2~\cite{oquab2024dinov2} to encode the static observation $o$ into global visual features $f_o$, and the cropped part observation ${o}^{part}$ into local features $f_{o^{part}}$. 
To ensure scale invariance, ${2Dbbox}^{{part}}$, ${2Dbbox}^{{base}}$ are normalized as $f_{2Dbbox^{part}}$, $f_{2Dbbox^{base}}$ by the corresponding resolution of the original image. The resulting reference part embedding $f^{ref}$ is defined as:
\begin{equation}
f^{\mathrm{ref}} = \{f_{o^{\mathrm{part}}},\, f_{\mathrm{2Dbbox}^{\mathrm{part}}},\, f_{\mathrm{2Dbbox}^{\mathrm{base}}}\}.
\label{eq:fref}
\end{equation}
Consequently, a training sample is parameterized by the tuple $(o', a, o, p^{ref})$, which simplifies to $(o', \emptyset, o, p^{ref})$ during the co-training phase to incorporate unannotated visual dynamics.
\begin{figure}[t] 
  \centering
  \includegraphics[width=\textwidth,trim=0cm 0cm 9cm 0cm, clip]{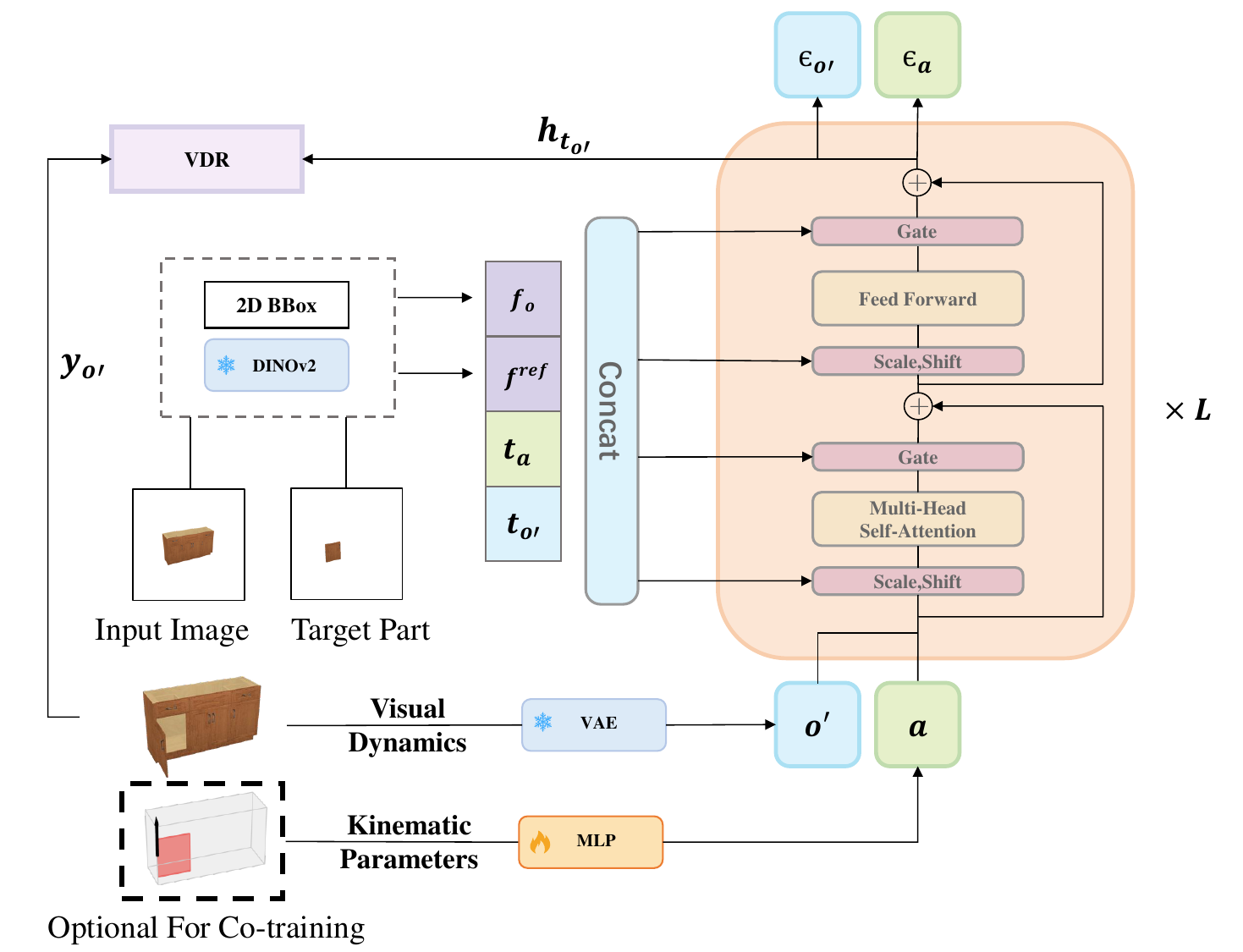}
  \caption{A single Part World Model (PWM-ArtGen) block consists of a transformer block with observation, a target articulated part, and diffusion timesteps conditioning via adaptive layer norm.}
  \label{fig:train}
\end{figure}
\subsection{Model Architecture}
\label{sec:Model}
We instantiate PWM-ArtGen as a diffusion Transformer that jointly models dynamic visual tokens and kinematic attributes, as shown in~\cref{fig:train}. The model predicts the actions and observation noises $\epsilon_a$ and $\epsilon_{o'}$, conditioned on static observations $o$ and the reference part $p^{ref}$, where $o$ is the input image and $p^{ref}$ specifies the target articulated part.
In accordance with the representation established in ~\cref{sec:REPRESENTATION}, the static observation $o$ is encoded as global visual features $f_o$, and the reference part $p^{ref}$ is embedded as $f^{ref}$. 
The concatenation of $f^{ref}$ with independent diffusion timesteps $t_a$ and $t_{o'}$ serves as the conditioning signal for the transformer blocks via adaptive layer normalization (AdaLN). For the visual branch, we operate in a latent image space by a frozen SDXL VAE and patchify the latent into $N$ tokens of width $D$, yielding $o'\in\mathbb{R}^{N\times D}$. Concurrently, for the kinematic branch, the kinematic attributes are projected into a tokenized space via a trainable MLP, denoted as $a\in\mathbb{R}^{20\times D}$ and concatenated with the visual tokens to form the input.

\subsection{Visual Dynamics Regularizer}
Inspired by REPA~\cite{yu2025repa}, we introduce a visual dynamics regularizer (VDR) to promote faithful part-level dynamics and mitigate reliance on static single-image evidence. Under joint training of action and visual branches, VDR biases the visual representations toward dynamics-consistent features that reflect action-induced motion. During training, the diffusion model produces joint hidden states $h_{t_{o'}}, h_{t_a} = f_\theta(z_t)$ at timestep $t$, where $z_t = (a_{t_a}, o'_{t_{o'}})$ and $h_{t_{o'}}^{[n]}$ denotes the feature of the $n$-th spatial patch in the visual branch. To ensure semantic consistency between the generated visual patches and the pretrained visual encoder, we apply the visual dynamics regularizer loss defined as

\begin{equation}
\mathcal{L}_{\mathrm{VDR}}(\theta)
:= -\,\mathbb{E}_{o',\, \epsilon_{o'},\, t_{o'},\, t_a}
\left[
\frac{1}{N} \sum_{n=1}^{N}
\operatorname{sim}\big(y_{o'}^{[n]},\, h_{t_{o'}}^{[n]}\big)
\right],
\label{eq:vdr_loss}
\end{equation}
where $y_{o'}^{[n]} = f(x_{o'})$ denotes the $n$-th patch feature 
of the clean image, and $\operatorname{sim}(\cdot,\cdot)$ 
is the cosine similarity function. This aligns the denoising trajectory with the semantic manifold of clean data, promoting robustness to noise while preserving essential visual structure.
The network in~\cref{fig:train} consists of $L$ layers in total, and only the visual hidden states from the first $l$ layers are aligned with the DINOv2 patch-wise features of the post-action observation $o'$ as shown in~\cref{eq:vdr_loss}.
\subsection{Co-Training}
\label{sec:Training}

PWM-ArtGen learns a joint diffusion process over kinematic parameters $a$ and post-action dynamic observation $o'$ conditioned on static observation $o$ and target articulated part $p^{\mathrm{ref}}$, enabling the model to incorporate part-level structural guidance during both training and inference. 
Specifically, we train a coupled noise prediction network~\cite{ho2020denoisingdiffusionprobabilisticmodels} 
$s_\theta(o, p^{\mathrm{ref}}, a_{t_a}, o'_{t_{o'}}, t_a, t_{o'})$ 
that approximates the conditional expectation of action and observation noise. 

To leverage both action-annotated and action-free data, we adopt a co-training strategy. Let $\mathcal{D}_{\mathrm{paired}}$ denote the dataset with ground-truth actions and $\mathcal{D}_{\mathrm{free}}$ denote the dataset containing only observations. For action-free samples, we impute the missing actions with pure Gaussian noise by fixing the action diffusion timestep to $T$ (\ie, $a_T = \epsilon_a \sim \mathcal{N}(0, 1)$). The corresponding denoising objective is formulated as follows:

\begin{equation}
\begin{aligned}
\mathcal{L}_{\epsilon_a,\, \epsilon_{o'}}(\theta) =\,
&\mathbb{E}_{\substack{
(o,\, p^{\mathrm{ref}},\, a,\, o') \sim \mathcal{D}_{\mathrm{paired}} \\
t_a,\, t_{o'} \sim \mathcal{U}(0,T) \\
\epsilon_a,\, \epsilon_{o'} \sim \mathcal{N}(0,1)
}}
\Bigl[
w_a \Vert \epsilon_a^{\theta} - \epsilon_a \Vert_2^2
+ w_{o'} \Vert \epsilon_{o'}^{\theta} - \epsilon_{o'} \Vert_2^2
\Bigr] \\
+\, \gamma &\mathbb{E}_{\substack{
(o,\, p^{\mathrm{ref}},\, \emptyset,\, o') \sim \mathcal{D}_{\mathrm{free}} \\
t_a = T,\, t_{o'} \sim \mathcal{U}(0,T) \\
\epsilon_a,\, \epsilon_{o'} \sim \mathcal{N}(0,1)
}}
\Bigl[
w_a \Vert \epsilon_a^{\theta} - \epsilon_a \Vert_2^2
+ w_{o'} \Vert \epsilon_{o'}^{\theta} - \epsilon_{o'} \Vert_2^2
\Bigr],
\end{aligned}
\label{eq:joint_eps_loss}
\end{equation}

\noindent\textit{where}
\begin{align}
\epsilon_a^{\theta},\, \epsilon_{o'}^{\theta}
&= s_{\theta}(o,\, p^{\mathrm{ref}},\, a_{t_a},\, o'_{t_{o'}},\, t_a,\, t_{o'}),\\
a_{t_a} &= \sqrt{\bar{\alpha}_{t_a}}\, a
+ \sqrt{1 - \bar{\alpha}_{t_a}}\, \epsilon_a,\\
o'_{t_{o'}} &= \sqrt{\bar{\alpha}_{t_{o'}}}\, o'
+ \sqrt{1 - \bar{\alpha}_{t_{o'}}}\, \epsilon_{o'}.
\end{align}

Here, $w_a$ and $w_{o'}$ are weights chosen to trade off between the action prediction and image prediction objectives, and $\gamma$ is a hyperparameter that balances the contribution of the action-free data. For any time step $t \in \{1, \dots, T\}$, we define $\alpha_t = 1 - \beta_t$ and $\bar{\alpha}_t = \prod_{s=1}^t \alpha_s$, where $\{\beta_s\}_{s=1}^T$ is a fixed variance schedule as in DDPM~\cite{ho2020denoisingdiffusionprobabilisticmodels}. 
The timesteps $t_a$ and $t_{o'}$ are sampled independently, meaning $\bar{\alpha}_{t_a}$ and $\bar{\alpha}_{t_{o'}}$ correspond to distinct noise levels for the action and observation diffusion processes, respectively.
This conditioning allows our model to align motion dynamics and visual appearance with part-level spatial context, 
providing a structured prior for generating coherent articulated behaviors. 

The final training objective combines the co-training loss with the visual dynamics regularization (VDR):
\begin{equation}
\mathcal{L} = \mathcal{L}_{\epsilon_a,\, \epsilon_{o'}} + \lambda \mathcal{L}_{\mathrm{VDR}},
\label{eq:final_loss}
\end{equation}
where $\lambda$ balances VDR loss with diffusion reconstruction accuracy. 
This joint training encourages articulation-aware dynamics while maintaining semantic coherence 
in the generated visual space.

\subsection{Synthetic-Real Mixed Training Strategy}
\begin{figure}[htbp] 
  \centering
  \includegraphics[width=\textwidth, trim=0cm 10cm 2cm 0cm, clip]{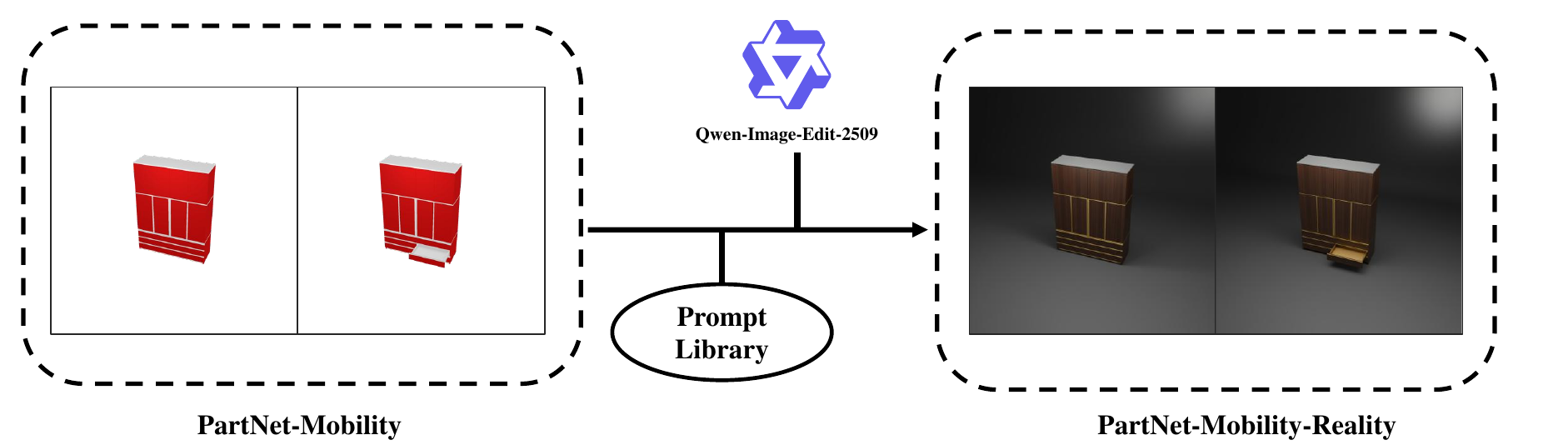} 
  \caption{Construction pipeline of PartNet-Mobility-Reality (PM-R). Synthetic renderings from PartNet-Mobility are enhanced into photorealistic counterparts through the Qwen-Image-Edit-2509~\cite{wu2025qwen} model guided by a structured Prompt Library, producing 19.7k photorealistic samples.}
  \label{fig:PM_R}
\end{figure}

To enhance the model's generalization ability on real-world images, we construct the PM-R dataset as illustrated in~\cref{fig:PM_R}. During co-training, we employ a synthetic–real mixed strategy using a length-proportional interleaving sampler that visits all samples once per epoch while mixing the two domains according to their relative sizes. At each epoch, the PartNet-Mobility-Reality and PartNet-Mobility subsets are independently shuffled, and a weighted round-robin procedure generates a global index order that maintains a consistent real–synthetic ratio. The sampler is DDP-compatible and partitions the sequence across workers without overlap, providing iteration-level balanced mixing. During training, the model jointly learns to predict both the action $a$ and the post-action observation $o'$, enabling a unified understanding of motion and visual dynamics. 

\section{Experiments}
\label{sec:Experiments}
We evaluate reconstruction quality and part connectivity graph prediction on both the ACD and PartNet-Mobility test sets. Qualitative results on ACD are conditioned on the ground-truth part graph. Furthermore, we comprehensively evaluate the visual dynamics of the image branch through both quantitative metrics and qualitative visualizations on the PartNet-Mobility test set. Ablation studies validate the contributions of key components: the image branch (IB) for joint modeling, the Visual Dynamics Regularizer (VDR), and co-training (Co-T) on PartNet-Mobility-Reality (PM-R) dataset. Additional qualitative results and experiments are provided in the Supplementary Material.

\subsection{Experimental Setup}
\noindent \textit{\textbf{Datasets.}} Following~\cite{Liu_ICLR_2025_SINGAPO}, we collect data from the PartNet-Mobility~\cite{Mo_CVPR_2019_PartNet} dataset to train our model, and from the ACD dataset~\cite{iliash2026s2o} for additional evaluation. 
Our training set consists of 473 articulated objects (985 movable parts in total) from PartNet-Mobility. 
Each object is rendered using \texttt{BLENDER\_EEVEE\_NEXT} under both the rest and open states, with 20 images per state sampled from random viewpoints within a $90^{\circ}$ horizontal and $90^{\circ}$ upward range relative to the object’s front. We extract part-level masks for the rest state and apply realistic augmentation strategies to construct PM-R. In total, we obtain about 19.7K rendered samples of PM for training and 19.7k PM-R image pairs without kinematic annotations for co-training. For testing, we use 77 unseen objects (155 parts) from PartNet-Mobility as the test split, each rendered from two random views in both states, yielding 310 test samples. To further evaluate the model’s generalization capability on realistic data, we include 135 objects (455 parts) from the ACD dataset in a zero-shot setting, producing 910 additional samples that better reflect real-world scenarios. To evaluate the visual branch, we render ten random views in both states from the PM test split, yielding 1550 test samples.
Please refer to the Supplementary Material for more details on rendering configurations.

\noindent \textit{\textbf{Compared methods.}} We select some representative methods: URDFormer~\cite{Chen_RSS_2024_URDFormer}, NAP~\cite{Lei_NeurIPS_2023_NAP}, SINGAPO~\cite{Liu_ICLR_2025_SINGAPO}, and Articulate-Anything~\cite{le2025articulate} for comparison of kinematic parameters; DragAPart~\cite{li2024dragapart}, and Puppet-Master~\cite{li2025puppet} for comparison of  visual dynamics. Due to the limited complexity of parts in the PartNet-Mobility dataset, these methods would suffer significant performance degradation, particularly in zero-shot scenarios, if not trained on larger datasets with more diverse part configurations, as reported in SINGAPO~\cite{Liu_ICLR_2025_SINGAPO}. In contrast, our method is specifically designed for part-level modeling and is thus largely unaffected by such dataset limitations. To ensure a fair comparison, we evaluated SINGAPO using its released checkpoint trained on 3,063 objects (approximately six times larger than PartNet-Mobility) and adopted the results of URDFormer and NAP-ICA reported in~\cite{Liu_ICLR_2025_SINGAPO}, where both methods were strictly adapted to the SINGAPO framework based on~\cite{Chen_RSS_2024_URDFormer, Lei_NeurIPS_2023_NAP}. Furthermore, we evaluate Articulate-Anything~\cite{le2025articulate} using GPT-4o~\cite{hurst2024gpt}, and directly utilize the officially released checkpoints for the visual dynamics baselines, DragAPart~\cite{li2024dragapart} and Puppet-Master~\cite{li2025puppet}.

\noindent \textit{\textbf{Metrics.}} To evaluate reconstruction quality and articulation correctness, we adopt four quantitative metrics:  
(1) ${d_{gIoU}~\downarrow}$, the generalized IoU between predicted and ground-truth part bounding boxes;  
(2) ${d_{cDist}~\downarrow}$, the Euclidean distance between the centers of corresponding parts;  
(3) ${d_{CD}~\downarrow}$, the Chamfer Distance between predicted and ground-truth meshes; and  (4) ${Acc~\uparrow}$, the accuracy of the predicted articulation graph.  
In addition, we report the average overlapping ratio ${AOR~\downarrow}$ introduced by CAGE~\cite{Liu_CVPR_2024_CAGE} to detect unrealistic collisions among sibling parts in the articulated graph.  
All metrics are computed over both resting and articulated states.  For clarity, we prefix the metric names with \textbf{RS-}(resting state) and \textbf{AS-}(articulated state) in the tables to indicate the corresponding evaluation state. We evaluate all methods under two settings: with and without access to the ground-truth part connectivity graphs and masks.
Following~\cite{li2024dragapart}, we evaluate the generated visual dynamics using standard image quality metrics, including pixel-level PSNR, patch-level SSIM, and feature-level LPIPS. We utilize ground-truth drag or part mask as a conditioning signal.
\begin{figure}[t] 
    \centering
    \includegraphics[width=\textwidth, keepaspectratio]{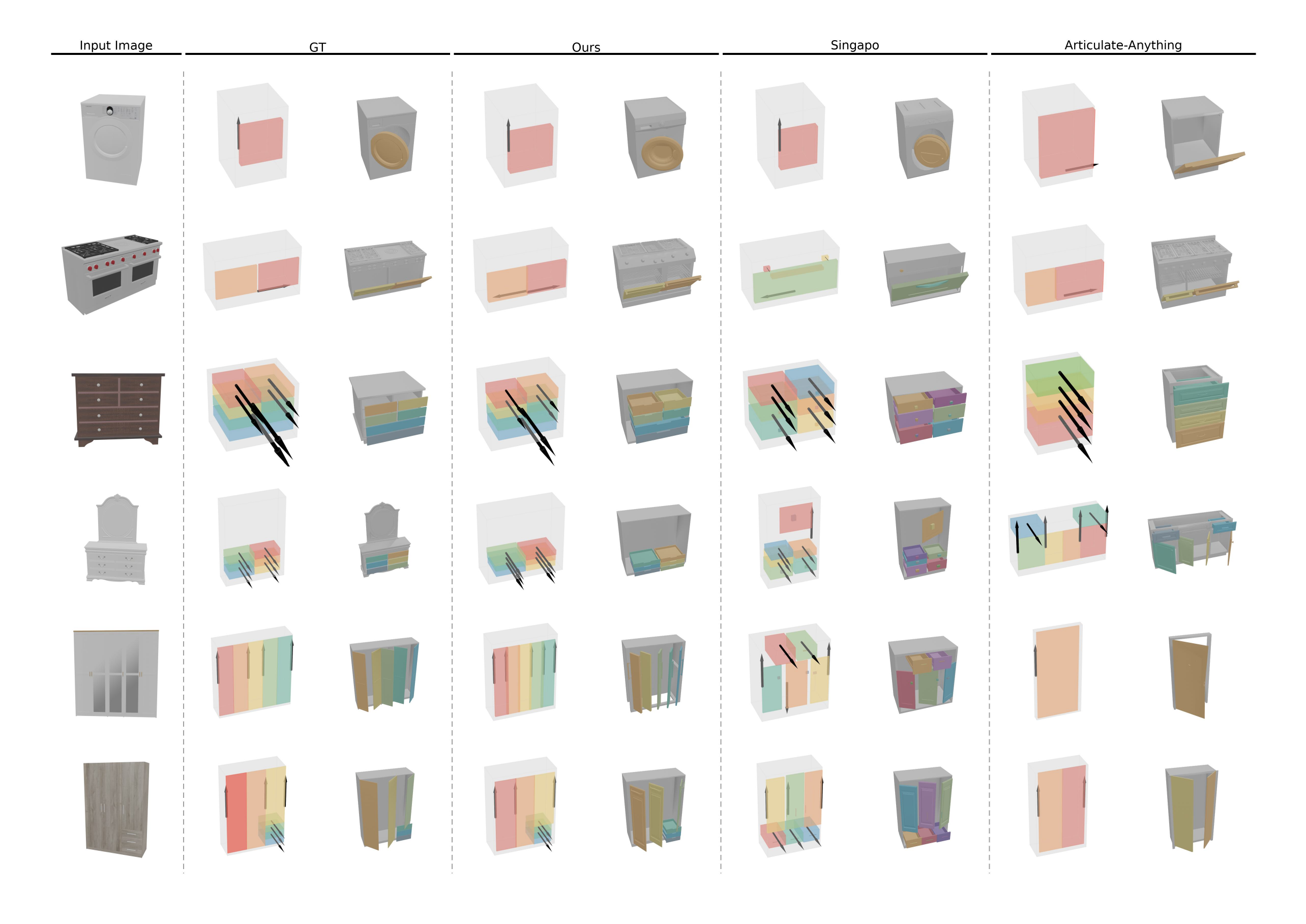}
    \caption{Qualitative comparison on the ACD dataset under zero-shot evaluation. From left to right: input image, ground truth (GT), our method, SINGAPO, and Articulate-Anything. Each row shows the predicted part layout and motion axes in both resting and articulated states. The last four rows highlight that other methods' predictions exhibit spatial misalignment with GT in part positioning and joint orientation. In contrast, our method produces parts that are \textbf{spatially better aligned with GT}, with more accurate joint orientations, cleaner boundaries, and more realistic articulation dynamics—especially on objects with complex textures and occlusions.}
    \label{fig:acd_test}
\end{figure}

\begin{table}[t] 
\centering
\caption{Comparison of reconstruction quality and graph prediction accuracy on the ACD test set.}
\label{tab:acd}

\resizebox{\textwidth}{!}{
\begin{tabular}{
    l
    *{6}{S[table-format=1.4]}
    S[table-format=1.4]  
    S[table-format=3.2]  
}
\toprule
& \multicolumn{6}{c}{Reconstruction quality} & \multicolumn{1}{c}{Collision} & \multicolumn{1}{c}{Graph}\\
\cmidrule(r){2-7} \cmidrule(l){8-9}
Method & {RS-$d_{\text{gIoU}}$ $\downarrow$} 
& {AS-$d_{\text{gIoU}}$ $\downarrow$} 
& {RS-$d_{\text{cDist}}$ $\downarrow$} 
& {AS-$d_{\text{cDist}}$ $\downarrow$} 
& {RS-$d_{\text{CD}}$ $\downarrow$} 
& {AS-$d_{\text{CD}}$ $\downarrow$} 
& {AOR$\downarrow$} 
& {Acc\%$\uparrow$} \\
\midrule
URDFormer-GTbbox   & 1.1986 & 1.2012 & 0.2292 & 0.2931 & 0.4343 & 0.4965 & 0.1209 & 48.53 \\
NAP-ICA-GTgraph    & 1.0585 & 1.0631 & 0.1987 & 0.2810 & 0.1376 & 0.2417 & 0.0193 & 36.67 \\
SINGAPO-GTgraph    & 0.9729 & 0.9767 & 0.1589 & 0.1968 & 0.1177 & 0.1752 & 0.0112 & 100.00 \\
\textbf{Ours-GTgraph} & $\mathbf{0.5693}$ & $\mathbf{0.5790}$ & $\mathbf{0.0687}$ & $\mathbf{0.1144}$ & $\mathbf{0.0569}$ & $\mathbf{0.0813}$ & $\mathbf{0.0031}$ & 100.00 \\
\midrule
URDFormer          & 1.2288 & 1.2309 & 0.2914 & 0.4285 & 0.7198 & 0.8995 & 0.2840 & 4.48 \\
NAP-ICA            & 1.0233 & 1.0286 & 0.1691 & 0.2331 & 0.1110 & 0.1887 & 0.0133 & 16.67 \\
SINGAPO    & 0.9775 & 0.9810 & 0.1574 & 0.2016 & 0.1100 & 0.1744 & 0.0103 & 39.34 \\
Articulate-Anything & 0.7574 & 0.7649 &  0.1938 & 0.2684 & 0.1826 & 0.2861 & $\mathbf{0.0039}$ & 40.00 \\
\textbf{Ours}      & $\mathbf{0.6822}$ & $\mathbf{0.6914}$ & $\mathbf{0.1480}$ & $\mathbf{0.1876}$ & $\mathbf{0.1033}$ & $\mathbf{0.1385}$ & 0.0906 & 39.34 \\
\bottomrule
\end{tabular}%
} 
\end{table}

\begin{table}[htbp] 
\centering
\caption{Comparison of reconstruction quality and graph prediction accuracy on the PartNet-Mobility test set.}
\label{tab:pm}
\resizebox{\textwidth}{!}{
\begin{tabular}{
    l
    *{6}{S[table-format=1.4]}
    S[table-format=1.3]
    S[table-format=4.2]
}
\toprule
& \multicolumn{6}{c}{Reconstruction quality} & \multicolumn{1}{c}{Collision} & \multicolumn{1}{c}{Graph}\\
\cmidrule(r){2-7} \cmidrule(l){8-9}
Method & {RS-$d_{\text{gIoU}}$ $\downarrow$} 
& {AS-$d_{\text{gIoU}}$ $\downarrow$} 
& {RS-$d_{\text{cDist}}$ $\downarrow$} 
& {AS-$d_{\text{cDist}}$ $\downarrow$} 
& {RS-$d_{\text{CD}}$ $\downarrow$} 
& {AS-$d_{\text{CD}}$ $\downarrow$} 
& {AOR$\downarrow$} 
& {Acc\%$\uparrow$} \\
\midrule
URDFormer-GTbbox  & 1.0861 & 1.0882 & 0.1471 & 0.3225 & 0.3400 & 0.6031 & 0.0616 & 83.55 \\
NAP-ICA-GTgraph    & 0.6830 & 0.6915 & 0.0739 & 0.2206 & 0.0282 & 0.2646 & 0.0105 & 44.81 \\
SINGAPO-GTgraph & 0.4345 & 0.4506 & $\mathbf{0.0319}$ & $\mathbf{0.0751}$ & $\mathbf{0.0108}$ & 0.0840 & 0.0019 & 100.00 \\
\textbf{Ours-GTgraph} & $\mathbf{0.3741}$ & $\mathbf{0.3803}$ & 0.0455 & 0.0913 & 0.0193 & $\mathbf{0.0678}$ & $\mathbf{0.0003}$ & 100.00 \\
\midrule
URDFormer          & 1.1868 & 1.1879 & 0.2693 & 0.4535 & 0.5502 & 0.8374 & 0.2341 & 32.03 \\
NAP-ICA            & 0.5778 & 0.5854 & 0.0501 & 0.0979 & $\mathbf{0.0173}$ & $\mathbf{0.0914}$ & 0.0120 & 75.97 \\
SINGAPO   & 0.4934 & 0.5065 & $\mathbf{0.0471}$ & $\mathbf{0.0968}$ & 0.0199 & 0.1107 & 0.0033 & 82.47 \\
Articulate-Anything & 0.4731 & 0.4870 &  0.1561 & 0.2302 & 0.1649 & 0.3126 & $\mathbf{0.0025}$ & 59.74 \\
\textbf{Ours}      & $\mathbf{0.4623}$ & $\mathbf{0.4707}$ & 0.0890 & 0.1415 & 0.0446 & 0.1272 & 0.0434 & 82.47 \\
\bottomrule
\end{tabular}%
}
\end{table}
\subsection{Implementation Details}
We train PWM-ArtGen for 200k steps on 4$\times$RTX 4090 GPUs with a global batch size of 128. The diffusion transformer features 12 layers, 12 attention heads, a hidden dimension of 768, an MLP ratio of 4, $2\times2$ patch embeddings, and 512-dimensional timestep embeddings. Optimization uses AdamW (learning rate $1\times10^{-4}$, weight decay $1\times10^{-6}$, $\beta=(0.9, 0.999)$, $\epsilon=1\times10^{-8}$). Inference relies on 10-step DDIM sampling~\cite{song2020denoising, ho2020denoisingdiffusionprobabilisticmodels}. For the training objectives, we set the action and image prediction weights to $w_a = w_{o'} = 1.0$, and the action-free data weight to $\gamma = 1.0$. To facilitate early visual-kinematic alignment, we apply the Visual Dynamics Regularizer (VDR) only during the first 80k training steps, with $\lambda = 0.5$, utilizing spatial features from a frozen DINOv2-Reg/14 encoder~\cite{oquab2024dinov2}. Additional details are in the Supplementary Material.

\subsection{Comparison with the State-of-the-art Methods}
\noindent \textit{\textbf{Quantitative Results On Kinematic.}} We report quantitative results on the ACD and PartNet-Mobility datasets in~\cref{tab:acd} and~\cref{tab:pm}, respectively. For fair comparison, we use the same part connection graph predicted by GPT-4o~\cite{hurst2024gpt}. All diffusion-based generative methods are evaluated five times per test sample, and the averaged results are reported to mitigate stochastic variations.

On the ACD test set, which contains more realistic and diverse articulated objects, our method achieves clearly superior performance across all metrics. Compared to existing baselines, PWM-ArtGen delivers the lowest reconstruction errors, demonstrating strong generalization capability to real-world-like data. Such consistent superiority on ACD highlights our model’s robustness under domain shift, which is a key challenge for articulated object generation.
\begin{figure}[t]
  \centering
  \includegraphics[width=0.8\textwidth, trim=5pt 220pt 400pt 10pt, clip]{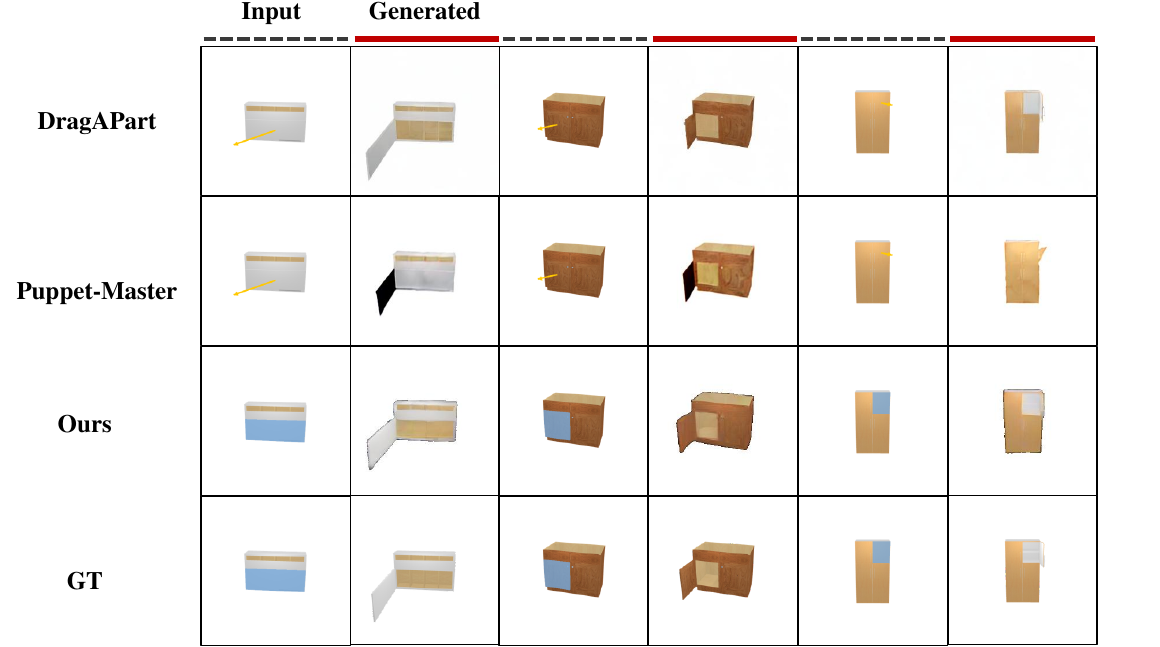}
  \caption{\textbf{Qualitative comparison of interactive dynamics generation.} 
    Compared to DragAPart and Puppet-Master, which utilize \textit{drag-based} prompts (yellow arrows), our model adopts \textit{mask-based} conditioning (blue regions). Black dashed lines and red solid lines denote the input and generated frames, respectively. Despite the significantly reduced parameter count, our method produces results consistent with the Ground Truth (GT) and competitive with state-of-the-art baselines.}
  \label{fig:dynamics-visual}
\end{figure}
\begin{table}[htbp]
  \centering
  \caption{Quantitative evaluation of the dynamics branch on the PM test set.}
  \label{tab:dynamics-results}
  \small 
  \setlength{\tabcolsep}{8pt}
  \begin{tabular}{@{}lccccc@{}}
    \hline
    Method & PSNR$\uparrow$ & SSIM$\uparrow$ & LPIPS$\downarrow$ & Parameters & Time\\
    \hline
    \textbf{DragAPart} & \textbf{24.803} & \textbf{0.943} & \textbf{0.079} & 1.43 B & 6.9 s\\
    Puppet-Master & 24.440 & 0.941 & 0.084 & 1.68 B & 11.6 s\\
    Ours & 24.328 & 0.906 & 0.093 & \textbf{0.38 B} & \textbf{0.7 s}\\
    \hline
  \end{tabular}
\end{table}

On the PartNet-Mobility test set, when the ground-truth graph and masks are provided, PWM-ArtGen performs on par or better than the strongest baseline across most metrics, with only marginal differences in a few cases. 
In the setting without ground-truth graph and masks, our method still outperforms competing approaches on several measures, though it exhibits slightly higher ${d_{cDist}}$ and ${d_{CD}}$ values. This minor degradation is mainly attributed to the current limitations of our Part Mask Generator, especially on synthetic datasets such as PM, where rendering artifacts and imperfect geometry annotations often lead to suboptimal mask quality.
In contrast, the ACD dataset provides relatively cleaner and more realistic object geometry, resulting in more reliable part-level supervision. It is worth noting that improving the segmentation quality of the Part Mask Generator is not the primary focus of this work, but this limitation can be effectively addressed in future research through more advanced segmentation pipelines. Overall, PWM-ArtGen exhibits strong reconstruction accuracy, reliable articulation prediction, and remarkable generalization to realistic data, which is one of the most challenging aspects of articulated generation.

\noindent \textit{\textbf{Qualitative Comparison On Kinematic.}}
As illustrated in~\cref{fig:acd_test}, PWM-ArtGen generates more consistent and physically plausible part motions than others. 
Our method preserves structural integrity during articulation and better aligns part boundaries, demonstrating improved understanding of 3D part relationships. The improvement is especially evident on the ACD dataset, where the objects exhibit complex materials and realistic lighting conditions, 
highlighting the strong generalization ability of PWM-ArtGen to real-world scenarios.

\noindent \textit{\textbf{Comparison On visual.}}
As illustrated in \cref{fig:dynamics-visual}, our model produces qualitative results comparable to Puppet-Master, while trailing slightly behind DragAPart. However, this gap is well-justified by the significant efficiency gains detailed in \cref{tab:dynamics-results}: our model achieves nearly 10$\times$ faster inference speed with only $0.25\times$ the parameters of DragAPart. These results indicate that our dynamics branch can generate visually plausible outputs. Furthermore, subsequent ablation studies demonstrate that explicit dynamics modeling significantly enhances the accuracy of kinematic estimation.
\subsection{Ablation Study}
\label{subsec:ablation}
We conduct detailed ablation studies to verify the effectiveness of each key component in \textbf{PWM-ArtGen}, including the Image Branch (IB) for joint modeling of visual dynamics and kinematics, the Visual Dynamics Regularizer (VDR), and the co-training (Co-T) on the PartNet-Mobility-Reality (PM-R) dataset. We construct several variants by selectively altering these components. The quantitative results are summarized in~\cref{tab:ablation} and~\cref{tab:ablation_pm}. 
\begin{table}[htbp]
\centering
\caption{\textbf{Ablation study on key design components for kinematic parameters.} Results are reported on the ACD dataset, assuming a ground truth (GT) part connectivity graph.}
\label{tab:ablation}
\resizebox{\textwidth}{!}{
\begin{tabular}{ccc|cccccc|c} 
\toprule
\multicolumn{3}{c|}{Settings} & \multicolumn{6}{c|}{Reconstruction Quality} & Collision \\
\cmidrule(lr){1-3} \cmidrule(lr){4-9} \cmidrule(lr){10-10} 
IB & VDR & Co-T & RS-$d_{gIoU}\!\downarrow$ & AS-$d_{gIoU}\!\downarrow$ & RS-$d_{cDist}\!\downarrow$ & AS-$d_{cDist}\!\downarrow$ & RS-$d_{CD}\!\downarrow$ & AS-$d_{CD}\!\downarrow$ & AOR$\!\downarrow$ \\
\midrule
 & & & 0.6567 & 0.6684 & 0.0801 & 0.1877 & 0.0796 & 0.1520 & 0.0252 \\ 
\midrule
\checkmark & & & 0.5948 & 0.6045 & 0.0739 & 0.1202 & 0.0681 & 0.0915 & 0.0033 \\
\checkmark & \checkmark & & 0.5885 & 0.5986 & 0.0699 & 0.1172 & 0.0587 & 0.0840 & 0.0038 \\
\checkmark & & \checkmark & 0.5786 & 0.5878 & 0.0696 & 0.1174 & 0.0619 & 0.0854 & \textbf{0.0020} \\
\midrule
\checkmark & \checkmark & \checkmark & \textbf{0.5693} & \textbf{0.5790} & \textbf{0.0687} & \textbf{0.1144} & \textbf{0.0569} & \textbf{0.0813} & 0.0031 \\
\bottomrule
\end{tabular}%
}
\end{table}

\begin{table}[htbp]
\centering
\caption{\textbf{Ablation study on key design components for visual dynamics.} Results are reported on the PM dataset.}
\label{tab:ablation_pm}
\resizebox{0.3\textheight}{!}{%
\begin{tabular}{ccc|ccc}
\toprule
\multicolumn{3}{c|}{Settings} & \multicolumn{3}{c}{Visual Dynamics Quality} \\
\cmidrule(lr){1-3} \cmidrule(lr){4-6}
IB & VDR & Co-T & PSNR$\uparrow$ & SSIM$\uparrow$ & LPIPS$\downarrow$ \\
\midrule
\checkmark & & & 21.964 & 0.835 & 0.132 \\
\checkmark & \checkmark & & 23.383 & 0.886 & 0.105 \\
\checkmark & & \checkmark & 22.908 & \textbf{0.908} & 0.107 \\
\midrule
\checkmark & \checkmark & \checkmark & \textbf{24.328} & 0.906 & \textbf{0.093} \\
\bottomrule
\end{tabular}%
}
\end{table}

\noindent \textit{\textbf{Effectiveness of Image Branch.}} We conduct ablation experiments to assess the contribution of jointly models visual dynamics and kinematic parameters. As shown in~\cref{tab:ablation}, modeling dynamics provides a strong inductive bias for kinematic estimation, leads to a noticeable degradation across all metrics rather than only action branch. Without the guidance of visual cues, the model struggles to infer motion directions and part boundaries accurately.

\noindent \textbf{The Impact of co-training.} 
Ablation results in~\cref{tab:ablation} and~\cref{tab:ablation_pm} demonstrate that omitting co-training on PM-R severely degrades performance under complex articulated states. Since PM-R closely aligns with real-world data distributions, its inclusion during co-training effectively bridges the synthetic-to-real gap. It enhances the model's ability to handle near-real inputs from ACD, leading to more precise kinematic estimation.

\noindent \textbf{Analysis of Visual Dynamics Regularizer.} 
As shown in~\cref{tab:ablation} and~\cref{tab:ablation_pm}, removing the VDR compromises part-wise coherence. This module imposes a critical inductive bias that encourages consistent motion and smoother surface deformations, thereby bridging the gap between static geometry and dynamic articulation.

\section{Conclusion}
In this work, we presented PWM-ArtGen, a unified Diffusion Transformer framework for generating articulated objects from a single image. By jointly modeling part-level visual dynamics and kinematic parameters through decoupled diffusion timesteps, our approach achieves a strong inductive bias for kinematic estimation without predefining part counts. Co-training on the PM-R dataset effectively leverages action-free visual priors and bridges the synthetic-to-real gap, while the Visual Dynamics Regularizer (VDR) ensures part-wise coherence and physically consistent motion. PWM-ArtGen achieves state-of-the-art performance on ACD and competitive results on PartNet-Mobility, demonstrating robust zero-shot generalization.

\section*{Acknowledgements}
This work was supported by the National Key Research and Development Program of China (2023YFA1008501).

\bibliographystyle{splncs04}
\bibliography{main}






\clearpage
\appendix

Our supplementary materials provide comprehensive implementation details and extended experimental validation for the proposed method, which can be summarized as follows:  
\begin{itemize}  
\item \textbf{Implementation details} of the Part Mask Generator pipeline, including SAM-based segmentation and GPT-4o connectivity graph prediction with visualized prompts.  
\item \textbf{Mathematical formulation} of the assembly strategy, including shared base bounding box initialization and 2D-to-3D positional guidance mechanism.  
\item \textbf{Retrieval pipeline adaptation} from SINGAPO with composite scoring metrics.  
\item \textbf{Complete training configurations} for both our model and the baseline, including hyperparameters, optimization settings, and computational resources.  
\item \textbf{Extended quantitative and qualitative results}, including hyperparameter analysis, baseline ablation studies on training data scale, robustness evaluations against imperfect structural priors, comparisons with the dual-state method, visual comparisons on PartNet-Mobility, and success/failure analyses on real-world ACD and in-the-wild or internet-sourced images. 
\end{itemize}
\section{Implementation Details}
\label{sec:rationale}

\subsection{Part Mask Generator}
\begin{figure}[t]
    \centering
    \includegraphics[width=\linewidth, trim=0cm 7cm 9cm 0cm, clip]{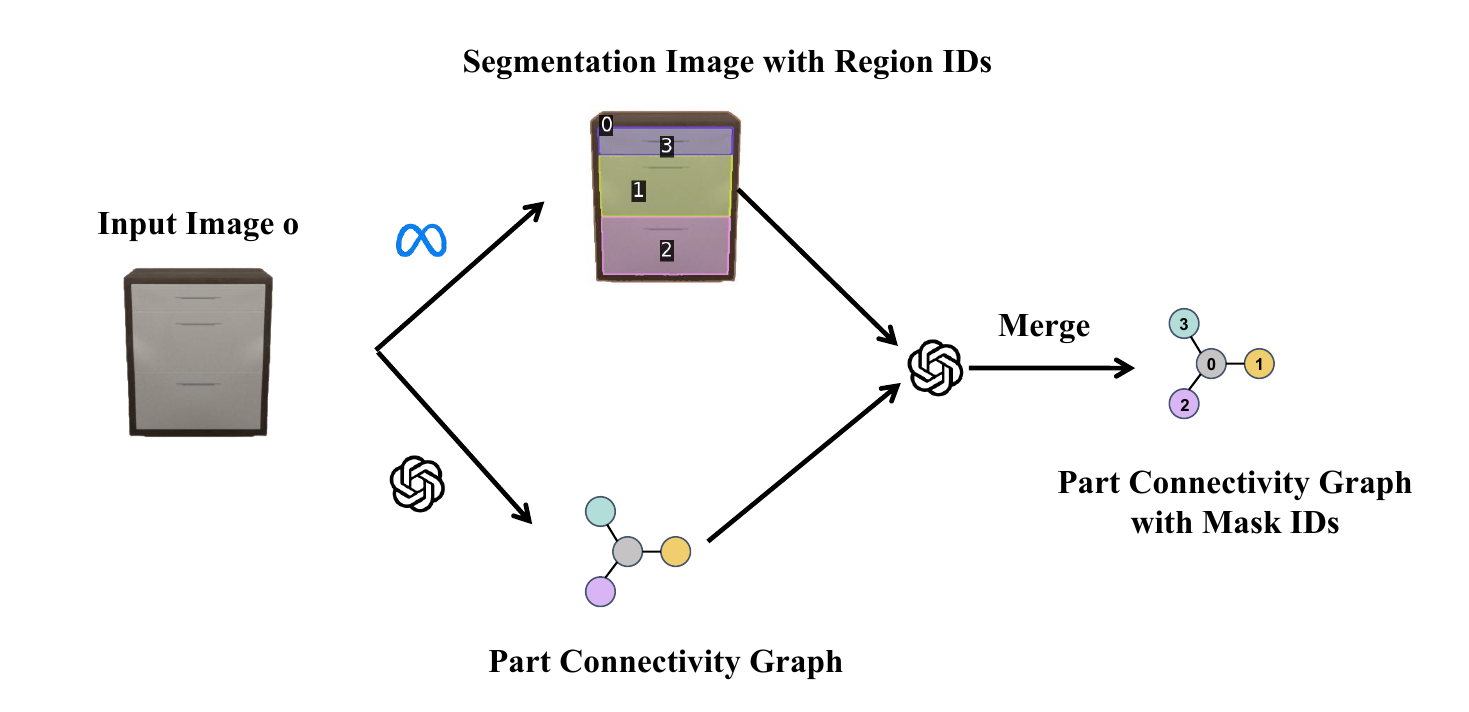}
    \caption{
        \textbf{Overview of the Part Mask Generator pipeline.} 
        Given an input image $o$, SAM generates a segmentation map with region IDs (top path), while GPT-4o independently infers an abstract part connectivity graph (bottom path). A final merging step aligns the two outputs, assigning mask IDs to graph nodes to produce the final Part Connectivity Graph with Mask IDs.
    }
    \label{fig:P_M_G}
\end{figure}

\noindent \textit{\textbf{Part Mask Segment using SAM.}} \\
We first apply the Segment Anything Model (SAM) to the input image $o$ in automatic mask generation mode, producing a set of candidate masks that serve as initial part proposals. To ensure semantic coherence, we merge masks with area below a threshold of 1000 pixels into their nearest larger neighbor based on centroid distance and IoU overlap. This yields a refined set of $N$ non-redundant part masks ${m_i}_{i=1}^N$, each representing a structural component of the object. An example output of this stage is shown in~\cref{fig:P_M_G}.\\

\begin{figure}[htbp]
    \centering
    \includegraphics[height=0.35\textheight, trim=0cm 4cm 5cm 0cm, clip]{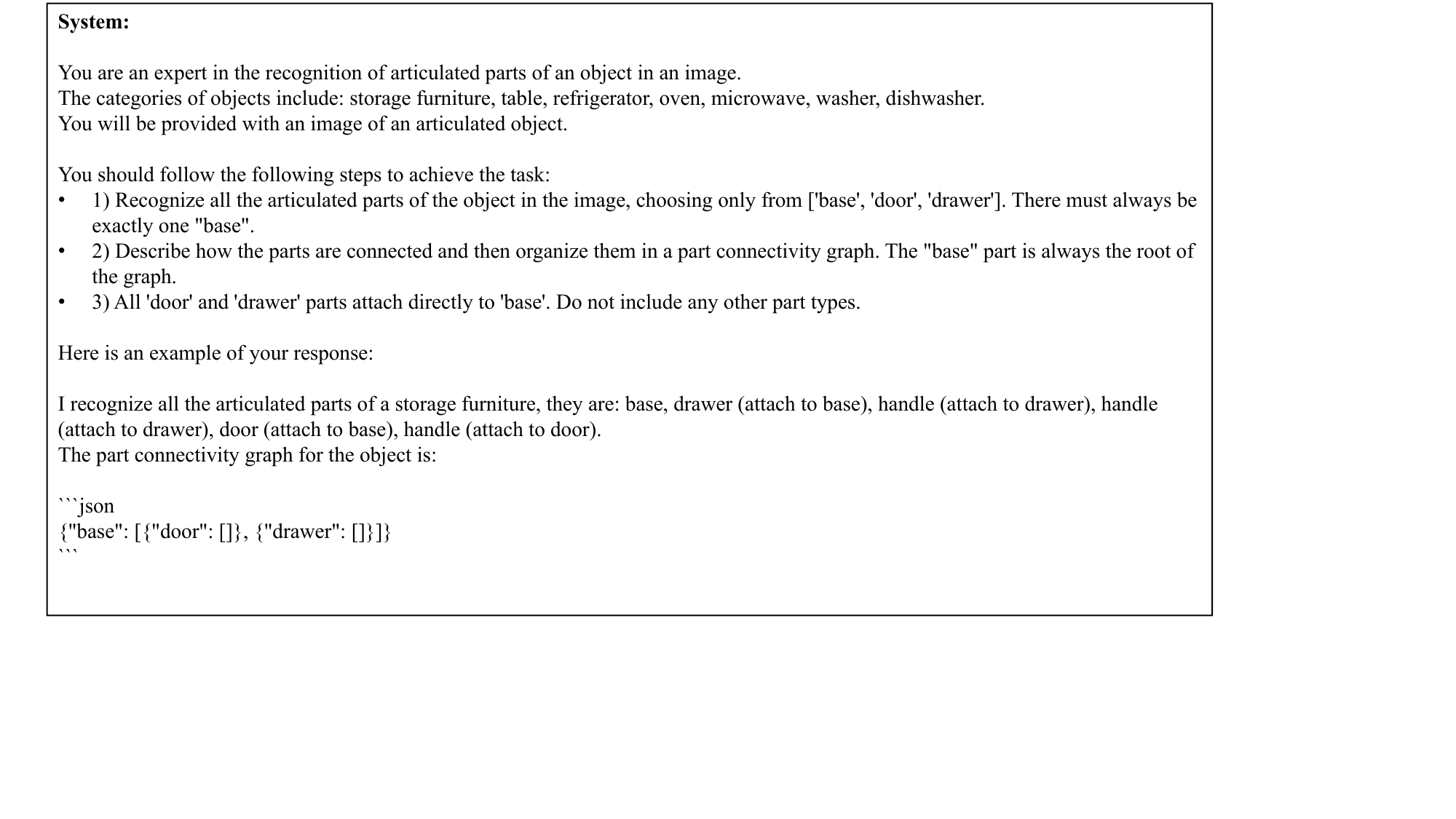}
    \caption{
        \textbf{First-step prompt for GPT-4o: Abstract graph generation.} 
        GPT-4o analyzes the original input image and outputs a natural language description of the object’s part structure and articulation relationships, without referencing mask IDs. This establishes the conceptual graph topology.
    }
    \label{fig:supp_prompt1}
\end{figure}

\begin{figure}[htbp]
    \centering
    \includegraphics[height=0.42\textheight, trim=0cm 1cm 5cm 0cm, clip]{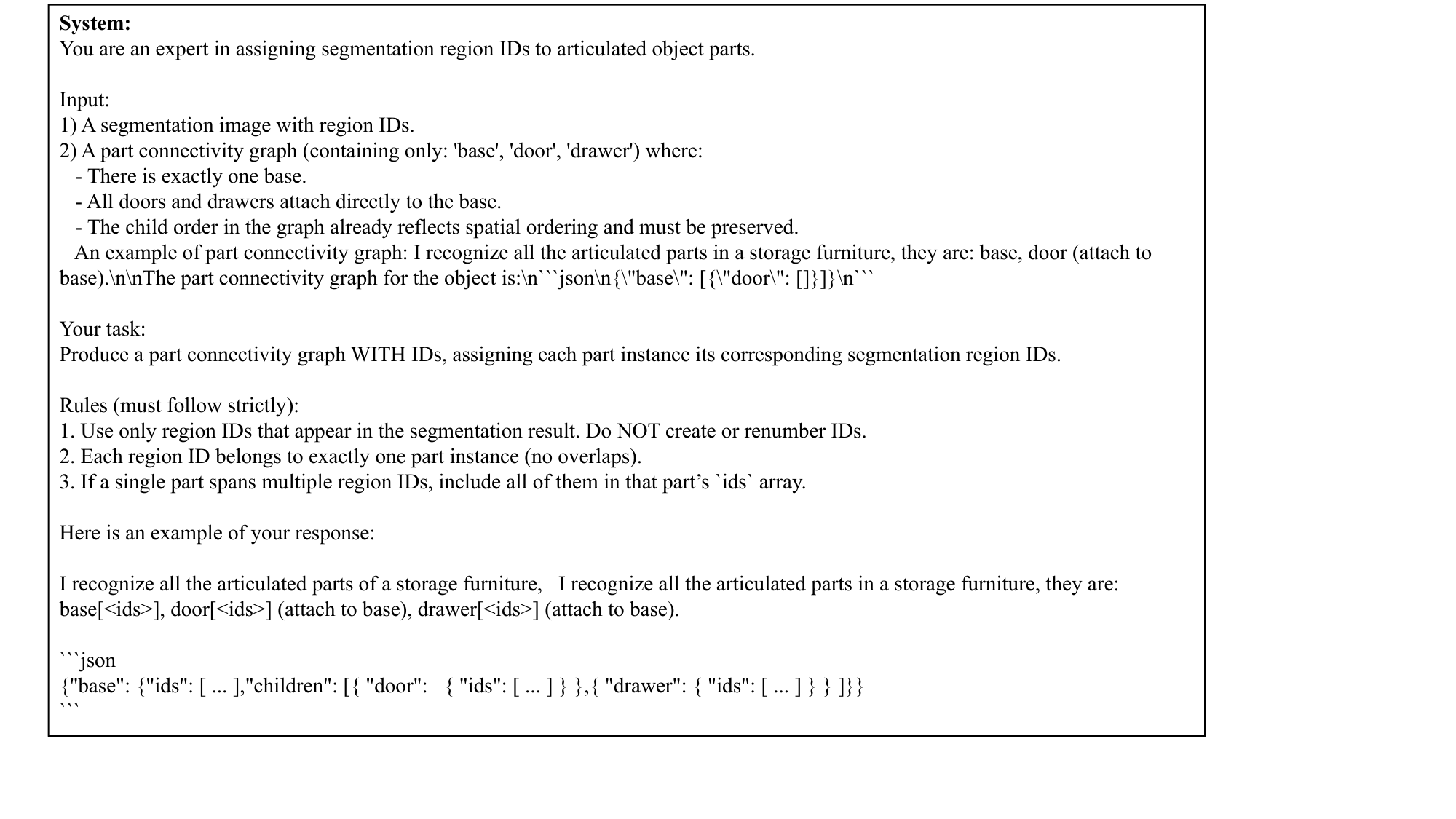}
    \caption{
        \textbf{Second-step prompt for GPT-4o: Mask ID assignment.} 
        The same image is overlaid with SAM-generated part masks, each labeled with a unique ID. GPT-4o maps the previously described parts to these IDs, producing a structured mapping that enables construction of the final \textit{Part Connectivity Graph with Assigned Mask IDs}.
    }
    \label{fig:supp_prompt2}
\end{figure}

\noindent \textit{\textbf{Part Connectivity Graph prediction.}} \\
We leverage GPT-4o (\texttt{gpt-4o-2024-08-06}) in a two-step reasoning process to construct an articulated connectivity graph from the segmented parts. In the first step, GPT-4o takes the original image $o$ as input and generates a textual description of the object’s part-level structure, identifying movable components and their joint types (e.g., door, drawer). In the second step, we overlay the image with the SAM-generated part masks, each labeled with a unique ID, and prompt GPT-4o to map the previously described parts to these IDs. The model outputs a node-to-mask correspondence, which we parse into a graph $G = (V, E)$. Unassigned graph nodes are discarded, while unassigned masks are merged into a single base node to preserve structural completeness. The prompt templates and model responses for both steps are visualized in~\cref{fig:supp_prompt1} and~\cref{fig:supp_prompt2}, respectively.
\subsection{Assembly}
Given the output of the Part Mask Generator, a part connectivity graph $G = (V, E)$ where each node $v_i$ is associated with a mask ID and classified. We assemble a coherent 3D articulated object using the \textbf{PWM-ArtGen} model. The assembly is anchored to a shared canonical coordinate frame defined by a base bounding box $\mathbf{b}^{base}$, ensuring spatial consistency across all parts.\\

\noindent \textbf{\textit{Shared Base Bounding Box Initialization.}} \\
We compute a shared base bounding box $\mathbf{b}^{base}= (\mathbf{c}^{{base}}, \mathbf{s}^{{base}})$ as follows:

Let $\mathcal{B}$ denote the set of indices corresponding to parts assigned to the base node. Collect the 3D bounding boxes $\{\textbf{$\mathbf{b}_i^{base}$}\}_{i \in \mathcal{B}}$ of these parts, each defined by center $\mathbf{c}_i^{base}$ and size $\mathbf{s}_i^{base}$, compute the average centroid and average size:
\begin{equation}
        \mathbf{c}^{base} = \frac{1}{|\mathcal{B}|} \sum_{i \in \mathcal{B}} \mathbf{c}_i^{base}, \quad
        \mathbf{s}^{base} = \frac{1}{|\mathcal{B}|} \sum_{i \in \mathcal{B}} \mathbf{s}_i^{base}.
\end{equation}
All parts are then transformed into the coordinate system defined by \textbf{$\mathbf{b}^{base}$}.\\
\noindent \textit{\textbf{2D-to-3D Positional Guidance for Part Alignment.}} \\
To ensure spatial consistency between the input image and the reconstructed 3D object, we align each part's 3D position according to its relative location within the base, using the base bounding box as the reference frame. Let the 2D base bounding box be defined by its center $\mathbf{c}^{\text{base,2D}} = (c^{\text{base,2D}}_x, c^{\text{base,2D}}_y)$ and size $\mathbf{s}^{\text{base,2D}} = (w^{\text{base,2D}}, h^{\text{base,2D}})$, where $w^{\text{base,2D}}$ and $h^{\text{base,2D}}$ are the width and height in pixel units. For part $i$, with 2D centroid $\mathbf{c}^{\text{part,2D}}_i = (c^{\text{part,2D}}_{i,x}, c^{\text{part,2D}}_{i,y})$, we compute its normalized coordinates within the base region:
\begin{equation}
    \begin{aligned}
        \hat{c}^{\text{2D}}_{i,x} &= \frac{c^{\text{part,2D}}_{i,x} - c^{\text{base,2D}}_x}{w^{\text{base,2D}}} + 0.5, \\
        \hat{c}^{\text{2D}}_{i,y} &= \frac{c^{\text{part,2D}}_{i,y} - c^{\text{base,2D}}_y}{h^{\text{base,2D}}} + 0.5,
    \end{aligned}
\end{equation}
where $\hat{\mathbf{c}}^{\text{2D}}_i = \big( \hat{c}^{\text{2D}}_{i,x},\, \hat{c}^{\text{2D}}_{i,y} \big) \in [0, 1]^2$.

In 3D space, the shared base bounding box is defined by its center $\mathbf{c}^{\text{base}} = (c^{\text{base}}_x, c^{\text{base}}_y, c^{\text{base}}_z) \in [-1, 1]^3$ and size $\mathbf{s}^{\text{base}} = (w^{\text{base}}, h^{\text{base}}, d^{\text{base}}) \in [0, 2]^3$, both expressed in the common normalized world coordinate system.. We interpret $\hat{\mathbf{c}}^{\text{2D}}_i$ as the desired relative position of part $i$ inside this base volume.

The 3D center $\mathbf{c}_i = (c_{i,x}, c_{i,y}, c_{i,z})$ of part $i$ is then updated to reflect this layout:
\begin{equation}
    \begin{aligned}
        c_{i,x} &\gets c^{\text{base}}_x + \left( \hat{c}^{\text{2D}}_{i,x} - 0.5 \right) w^{\text{base}}, \\
        c_{i,y} &\gets c^{\text{base}}_y + \left( \hat{c}^{\text{2D}}_{i,y} - 0.5 \right) h^{\text{base}},
    \end{aligned}
\end{equation}
while the Z-coordinate $c_{i,z}$ remains unchanged.

This mapping ensures that a part located at the left edge of the base in the 2D image is placed at the left boundary of the 3D base volume, and similarly for all other relative positions. The alignment is applied only when the initial assembly exhibits significant inter-part collisions (\eg collision ratio $> 10\%$); otherwise, the original positions are retained to avoid disrupting a physically plausible configuration. No rotation or scaling is applied—only the XY position is adjusted to reflect the 2D spatial layout within the 3D base reference frame.

\subsection{Retrieval}
We adapt the mesh retrieval pipeline from SINGAPO, by replacing its Abstract Instantiation Distance (AID) with a composite score combining AS/RS-$d_{\text{gIoU}}$ and AS/RS-$d_{\text{cDist}}$ to select the best-matching shape from the database.

\subsection{Training Details}
\noindent{\textit{\textbf{Our model training.}}}\\
Our model is trained for 200k steps on 4$\times$RTX 4090 GPUs with a global batch size of 128. The diffusion process is defined over 100 timesteps using the DDIM framework. We explicitly set $l=8$ for VDR. For evaluation, all inference experiments are conducted on a single RTX 4090 GPU. Furthermore, the construction of our PM-R dataset took approximately one week utilizing 4$\times$L40 GPUs.

\noindent{\textit{\textbf{SINGAPO training.}}}\\
We include the method from Singapo as a baseline, following the training configuration reported in their work. The model is initialized with CAGE-pretrained weights, which provides a strong prior over articulated object dynamics conditioned on scene graphs; the model then modulates this prior to better align with the input image. Training proceeds for 200 epochs with a batch size of 20, sampling 16 timesteps per iteration from a 1,000-step diffusion process. Optimization uses AdamW with differential learning rates: $5\times10^{-4}$ for the ICA module and $5\times10^{-5}$ for the base model, with $\beta = (0.9, 0.99)$. The learning rate is warmed up over 3 epochs from $1\times10^{-6}$ to its peak value, followed by cosine annealing down to $1\times10^{-5}$. The network employs 6 attention blocks, each with 4 heads and 128 hidden units. Training is conducted on 4 RTX 4090 GPUs and takes approximately 12 hours.

\section{Additional Experiments and Results}
\subsection{Hyperparameter analysis}
\begin{figure}[htbp]
    \centering
    \includegraphics[width=0.75\linewidth]{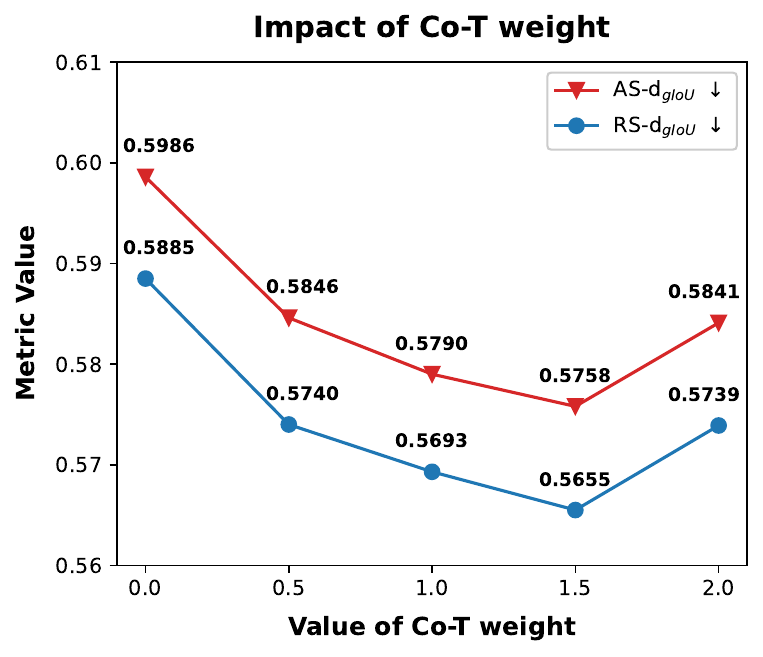}
    \caption{Sensitivity analysis of the Co-T weight on the ACD dataset.}
    \label{fig:cot_weight}
\end{figure}

We provide a sensitivity analysis of the co-training weight in~\cref{fig:cot_weight}. Taking RS-$d_{\mathrm{gIoU}}$ and AS-$d_{\mathrm{gIoU}}$ as representative metrics, the performance remains relatively stable across different weight values. In our experiments, we set the co-training weight to 1.0 as a balanced default configuration.
\subsection{Justification for Baseline Selection and Fair Comparison}
As shown in~\cref{tab:singapo_comparison}, the released checkpoint of Singapo trained on a much larger dataset of 3,063 objects significantly outperforms variants retrained on our limited PartNet-Mobility datasets. This performance gap confirms that Singapo, as originally proposed, is highly sensitive to training data scale and part configuration diversity. To ensure a fair comparison with our method designed for part-level modeling and robust to dataset scale, we adopt the released checkpoint as the canonical baseline. This avoids conflating architectural advantages with data-scale biases, aligning with the original evaluation protocol.
\begin{table}[t]
\centering
\caption{To ensure fair comparison, we evaluate Singapo’s released checkpoint (trained on 3,063 objects) against its variants retrained on PM, using the same evaluation protocol on PM and ACD test sets.}
\label{tab:singapo_comparison}
\resizebox{\textwidth}{!}{%
\begin{tabular}{
    l
    l
    *{6}{S[table-format=1.4]}
    S[table-format=1.4]
}
\toprule
\multicolumn{1}{c}{Testing data} & \multicolumn{1}{c}{Checkpoint} & \multicolumn{6}{c}{Reconstruction quality} & \multicolumn{1}{c}{Collision} \\
\cmidrule(r){3-8} \cmidrule(l){9-9}
& & {RS-$d_{\text{gIoU}}$ $\downarrow$} 
& {AS-$d_{\text{gIoU}}$ $\downarrow$} 
& {RS-$d_{\text{cDist}}$ $\downarrow$} 
& {AS-$d_{\text{cDist}}$ $\downarrow$} 
& {RS-$d_{\text{cD}}$ $\downarrow$} 
& {AS-$d_{\text{cD}}$ $\downarrow$} 
& {AOR$\downarrow$} \\
\midrule
\multirow{2}{*}{PM} 
& PM          & 0.5965 & 0.6101 & 0.0827 & 0.1427 & 0.0606 & 0.1725 & 0.0032 \\
& Released    & \textbf{0.4345} & \textbf{0.4506} & \textbf{0.0319} & \textbf{0.0751} & \textbf{0.0108} & \textbf{0.0840} & \textbf{0.0019} \\
\midrule
\multirow{2}{*}{ACD} 
& PM          & 1.1270 & 1.1306 & 0.2260 & 0.2782 & 0.1948 & 0.2783 & 0.0156 \\
& Released   & \textbf{0.9729} & \textbf{0.9767} & \textbf{0.1589} & \textbf{0.1968} & \textbf{0.1177} & \textbf{0.1752} & \textbf{0.0112} \\
\bottomrule
\end{tabular}
}
\end{table}

\subsection{Robustness to Imperfect Part Masks and Graph Predictions}
To evaluate the robustness of PWM-ArtGen against imperfect structural priors, we quantify performance degradation under varying levels of noise and incomplete data (see~\cref{tab:singapo_comparison_reduced}). 
\begin{table}[t]
\centering
\caption{
    Quantitative evaluation of model robustness against imperfect structural priors on the PM and ACD test sets. We compare reconstruction quality using perfect Ground Truth (GT, $d=0$), systematically degraded GT with $d$ randomly dropped movable parts, and noisy predictions from a fully automated SAM + GPT-4o pipeline. The variable $n$ denotes the average number of parts remaining per object. The results demonstrate that our method degrades gracefully under severe part omissions and maintains strong performance when handling real-world prediction noise. All results are averaged over 5 samples.
}
\label{tab:singapo_comparison_reduced}
\resizebox{0.8\textwidth}{!}{
\begin{tabular}{
    l
    l
    *{4}{S[table-format=1.4]}
}
\toprule
\multicolumn{1}{c}{Testing data} & \multicolumn{1}{c}{Setting} & \multicolumn{4}{c}{Reconstruction quality} \\
\cmidrule(lr){3-6}
& & {RS-$d_{\text{cDist}}$ $\downarrow$} 
& {AS-$d_{\text{cDist}}$ $\downarrow$} 
& {RS-$d_{\text{cD}}$ $\downarrow$} 
& {AS-$d_{\text{cD}}$ $\downarrow$} \\
\midrule

\multirow{6}{*}{PM}
& \textbf{GT (d=0, n=3.01)} & \textbf{0.0455} & \textbf{0.0913} & \textbf{0.0193} & \textbf{0.0678} \\
& GT (d=1, n=2.01)    & 0.1072 & 0.1582 & 0.0754 & 0.1460 \\
& GT (d=2, n=1.45)    & 0.1364 & 0.1985 & 0.1035 & 0.1923 \\
& GT (d=3, n=1.21)    & 0.1477 & 0.2129 & 0.1142 & 0.2078 \\ 
& GT (d=4, n=1.05)    & 0.1541 & 0.2210 & 0.1224 & 0.2178 \\ 
& SAM + GPT-4o (n=2.94)          & 0.0890 & 0.1415 & 0.0446 & 0.1272 \\
\midrule

\multirow{6}{*}{ACD}
& \textbf{GT (d=0, n=4.40)} & \textbf{0.0687} & \textbf{0.1144} & \textbf{0.0569} & \textbf{0.0813} \\
& GT (d=1, n=3.40)   & 0.1346 & 0.1780 & 0.1182 & 0.1605 \\
& GT (d=2, n=2.58)   & 0.1477 & 0.1903 & 0.1288 & 0.1754 \\
& GT (d=3, n=2.06)   & 0.1544 & 0.1967 & 0.1413 & 0.1916 \\
& GT (d=4, n=1.69)   & 0.1563 & 0.1992 & 0.1518 & 0.2042 \\
& SAM + GPT-4o (n=3.29)        & 0.1480 & 0.1876 & 0.1033 & 0.1385 \\
\bottomrule

\end{tabular}%
}
\end{table}

\noindent \textbf{\textit{Experimental Setup.}}\\
We simulate incomplete graphs and masks by randomly dropping up to $d$ movable parts ($d \in \{1, 2, 3, 4\}$) and their corresponding masks from the Ground Truth (GT) annotations, stopping when no movable parts remain. Note that $d=0$ denotes the perfect GT baseline. The variable $n$ indicates the average number of remaining parts (base + movable) per object. To assess real-world applicability, we also evaluate the performance of a fully automated pipeline utilizing predictions from SAM and GPT-4o.

\noindent \textbf{\textit{Metric Selection for Robustness Analysis.}}\\
In~\cref{tab:singapo_comparison_reduced}, we exclusively report $d_{\text{cDist}}$ and $d_{\text{cD}}$ to ensure mathematically consistent evaluations across varying degrees of structural corruption $d$. Other metrics are purposefully omitted due to severe confounding factors when comparing across different part counts:  (1) The collision metric (AOR) becomes paradoxically skewed: dropping more parts inherently reduces collision probability, artificially improving the score. (2) Volume-dependent metrics ($d_{\text{gIoU}}$) suffer from baseline volume shifts when parts are removed, rendering cross-$d$ degradation trends ill-posed. Conversely, point-wise distances ($d_{\text{cDist}}$ and $d_{\text{cD}}$) provide stable, unbiased measurements of spatial localization and geometric fidelity for the \textit{remaining} parts, independent of the overall graph size.

\noindent \textbf{\textit{Tolerance to Missing Parts.}} \\
As anticipated, increasing the number of dropped parts ($d=1 \to 4$) results in a gradual decline in reconstruction quality across both the PM and ACD datasets. Crucially, this degradation is graceful rather than catastrophic. Even under extreme information loss, such as $d=4$ on the PM dataset where the average part count $n$ drops precipitously from 3.01 to 1.05, our model sustains reasonable reconstruction fidelity, evidencing strong fault tolerance to incomplete structural priors.

\noindent \textbf{\textit{Robustness to Automated Pipeline Noise.}} \\
When evaluated on the inputs generated by SAM + GPT-4o, PWM-ArtGen demonstrates highly competitive practical robustness across all evaluated metrics. On the PM dataset, the automated pipeline inevitably introduces a performance drop from the pristine GT baseline ($d=0$). However, it consistently outperforms the scenario where a single GT part is omitted ($d=1$) across both $d_{\text{cDist}}$ and $d_{\text{cD}}$ metrics. For the structurally more complex ACD dataset, the automated pipeline's performance generally falls within the degradation bounds of missing one to two parts ($d=1$ and $d=2$). Specifically, it surpasses the $d=1$ setting in Chamfer Distance metrics ($\text{RS-}d_{\text{cD}}$: $0.1033$ vs.\ $0.1182$), while aligning more closely with the $d=2$ setting on center distances ($\text{RS-}d_{\text{cDist}}$: $0.1480$ vs.\ $0.1477$). These results validate that our model effectively tolerates real-world prediction noise, maintaining generation quality comparable to missing only 1--2 structural parts.

These findings confirm that PWM-ArtGen does not strictly overfit to perfect annotations. Instead, it effectively synthesizes articulated structures even when conditioned on severe part omissions and noisy masks from off-the-shelf vision-language models.

\subsection{Comparison with Dual-State Method}
To further contextualize our single-image approach, we compare it against DIPO, a dual-state method referenced in our related work. By utilizing both rest and open state images, DIPO accesses explicit visual cues of the articulation process, thus serving as an information-rich upper bound.

Due to the lack of released data processing scripts, fully reproducing DIPO's exact experimental conditions is practically difficult. However, relying on their reported metrics, our single-image approach achieves a graph accuracy highly comparable to DIPO's explicit dual-state framework. This comparable level of topological completeness provides a fair, mathematically sound, and highly referential basis for evaluating volume-dependent metrics like $d_{\text{gIoU}}$.

Under this equitable baseline, our method demonstrates highly competitive bounding box localization on the PM dataset despite lacking the second observation state. Furthermore, on the challenging out-of-domain ACD dataset, our method significantly outperforms DIPO in bounding box metrics (RS/AS-$d_{\text{gIoU}}$). We attribute this to our co-training strategy and Visual Dynamics Regularizer (VDR), which endow our model with robust generalization capabilities, mitigating the domain gap issues inherent in acquiring perfectly consistent multi-state real-world pairs. While DIPO retains an expected advantage in fine-grained distance metrics ($d_{\text{cDist}}$, $d_{\text{cD}}$) due to its dual-view geometric constraints, our findings confirm that single-image generation can robustly approximate, and in terms of real-world localization, even surpass multi-state methods.

\begin{table}[t]
\centering
\caption{\textbf{Comparison with the dual-state method DIPO.} While DIPO requires both rest and open state images as inputs (serving as an information-rich upper bound), our method operates strictly on a single image.}
\label{tab:dipo_comparison}
\resizebox{\textwidth}{!}{%
\begin{tabular}{
    l
    l
    *{6}{S[table-format=1.4]}
    S[table-format=2.2] 
}
\toprule
\multicolumn{1}{c}{Testing data} & \multicolumn{1}{c}{Method} & \multicolumn{6}{c}{Reconstruction quality} & \multicolumn{1}{c}{Graph} \\
\cmidrule(r){3-8} \cmidrule(l){9-9}
& & {RS-$d_{\text{gIoU}}$ $\downarrow$} 
& {AS-$d_{\text{gIoU}}$ $\downarrow$} 
& {RS-$d_{\text{cDist}}$ $\downarrow$} 
& {AS-$d_{\text{cDist}}$ $\downarrow$} 
& {RS-$d_{\text{cD}}$ $\downarrow$} 
& {AS-$d_{\text{cD}}$ $\downarrow$} 
& {Acc\%$\uparrow$} \\
\midrule
\multirow{2}{*}{PM} 
& DIPO          & \textbf{0.4561} & \textbf{0.4683} & \textbf{0.0359} & \textbf{0.0732} & \textbf{0.0132} & \textbf{0.0423} &  85.06 \\
& Ours    & 0.4623 & 0.4707 & 0.0890 & 0.1415 & 0.0446 & 0.1272 &  82.47 \\
\midrule
\multirow{2}{*}{ACD} 
& DIPO          & 0.9126 & 0.9151 & \textbf{0.1253} & \textbf{0.1541} & \textbf{0.0751} & \textbf{0.1085} &  48.15 \\
& Ours   & \textbf{0.6822} & \textbf{0.6914} & 0.1480 & 0.1876 & 0.1033 & 0.1385 &  39.34 \\
\bottomrule
\end{tabular}%
}
\end{table}
\subsection{Qualitative Evaluation Across Diverse Datasets}
\begin{figure}[t]
    \centering
    \includegraphics[width=\textwidth, trim=1cm 1cm 1cm 1cm, clip]{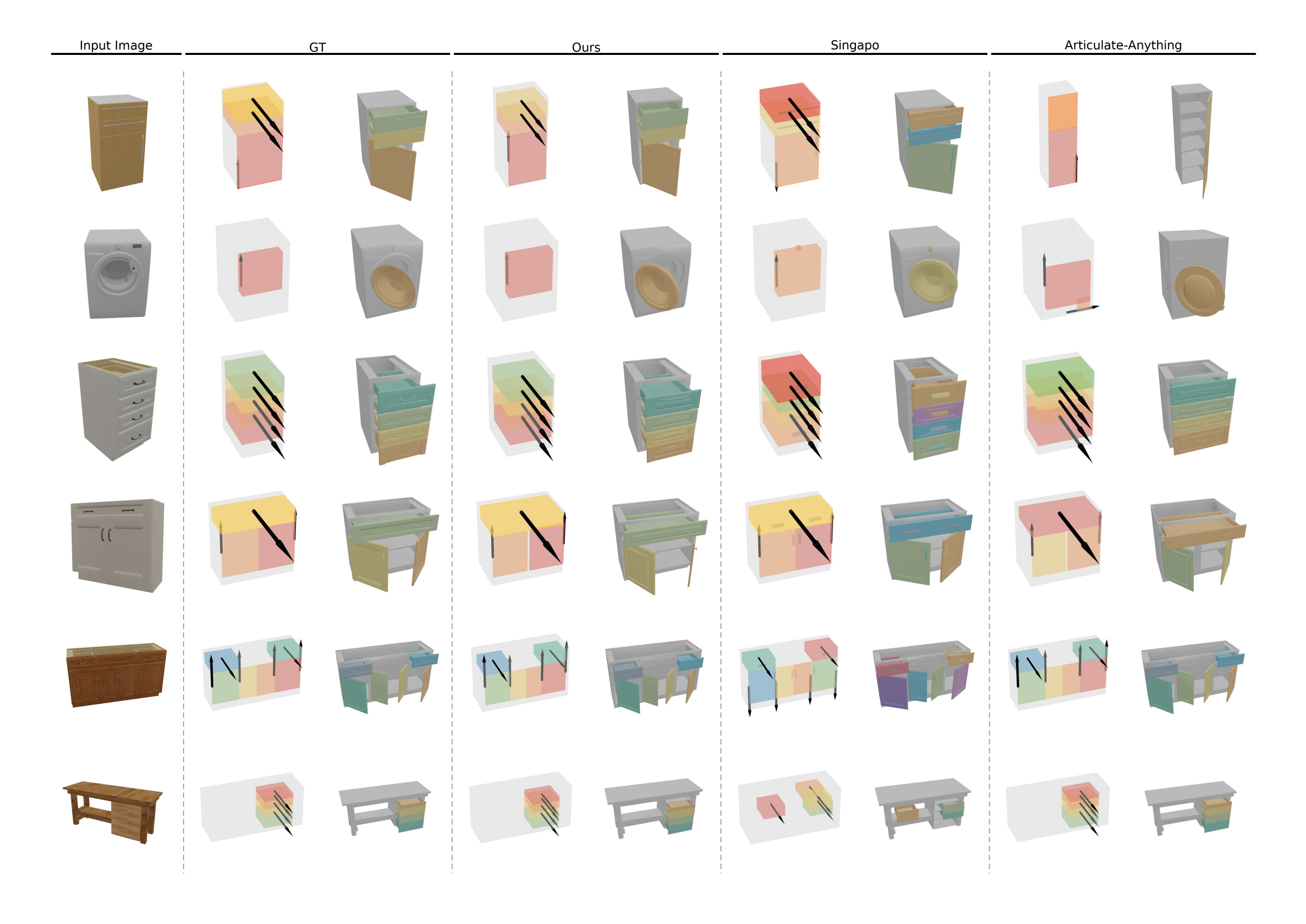}
    \caption{Qualitative comparison on the PartNet-Mobility dataset. From left to right: input image, ground truth (GT), our method, SINGAPO, and Articulate-Anything. Each row shows the predicted part layout and motion axes in both resting and articulated states. The last row highlights that SINGAPO’s predictions exhibit spatial misalignment with GT in part positioning and joint orientation. In contrast, our method produces parts that are \textbf{spatially better aligned with GT}, with more accurate joint orientations, cleaner boundaries, and more realistic articulation dynamics—especially on objects with complex textures and occlusions.}
    \label{fig:pm_test}
\end{figure}

We have presented qualitative comparisons on the ACD dataset, demonstrating our method's superior generalization under complex materials and realistic lighting. We now provide complementary results on the PartNet-Mobility test set in~\cref{fig:pm_test} to further validate our method’s robustness in controlled, synthetic environments, where part geometry and articulation are structured, yet occlusion and texture variation pose challenges.

\begin{figure}[htbp]
    \centering
    \includegraphics[width=\textwidth, trim=0cm 0cm 0cm 0cm, clip]{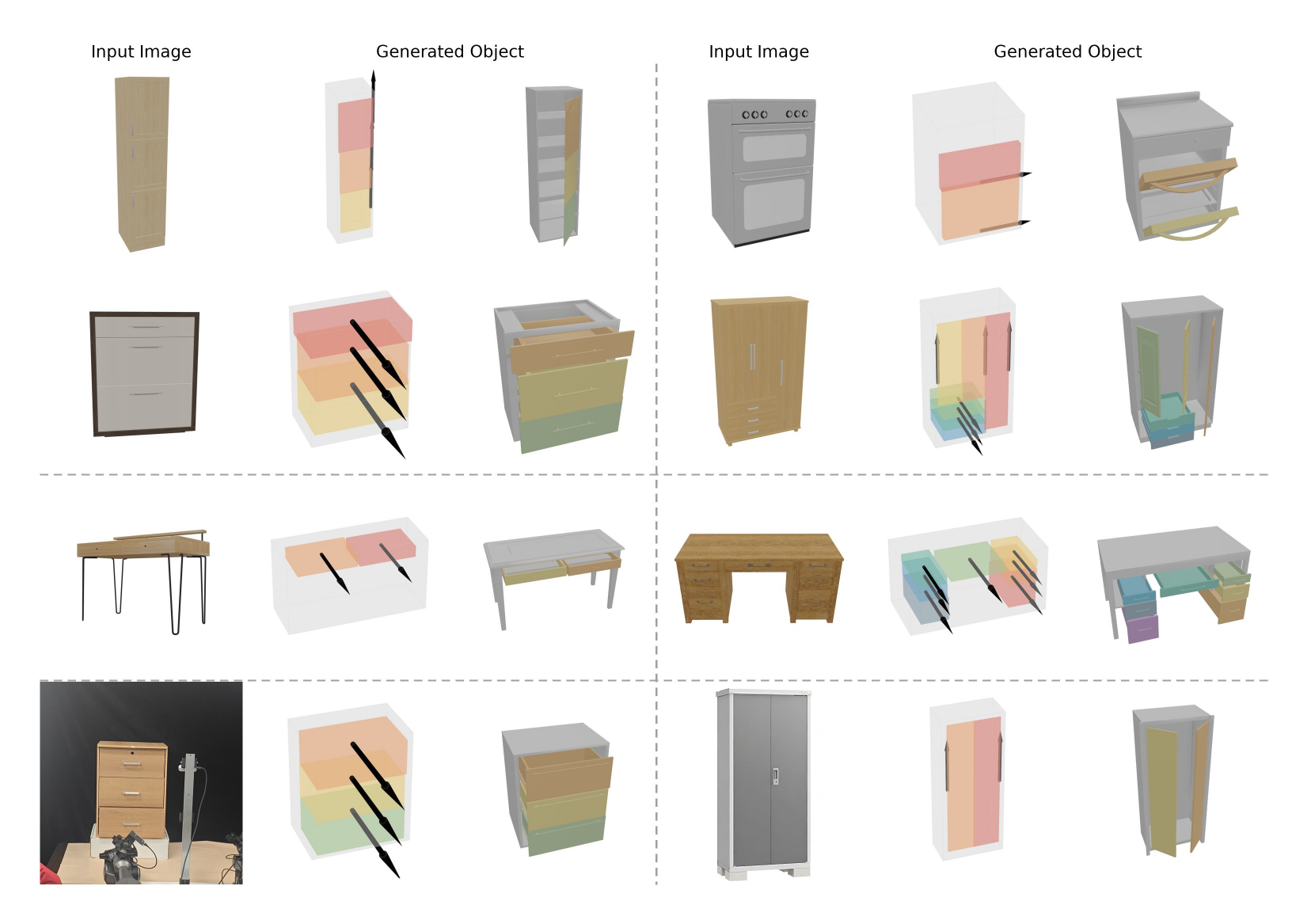}
    \caption{
        Additional qualitative results on the ACD test dataset and in-the-wild or internet-sourced images.  
    \textbf{First two rows:} Successful reconstructions from ACD test inputs, demonstrating robust part-level articulation recovery under complex textures, occlusions, and lighting variations.  
    \textbf{Third row:} Failure cases, typically occurring when input objects exhibit base geometries or articulation patterns outside the training distribution (\eg non-standard supports, irregular hinge placements), resulting in physically implausible or collision-prone configurations due to mismatched 3D priors.  
    \textbf{Last row:} Zero-shot generalization on in-the-wild or internet-sourced images, showcasing the model’s ability to infer articulated structures from real-world photos beyond curated datasets.
    }
    \label{fig:more_test}
\end{figure}

~\cref{fig:more_test} shows additional results across three categories. \textbf{The first two rows} highlight successful generations from ACD inputs, demonstrating robust recovery of articulated structures under occlusion and varied materials. \textbf{The third row} presents failures, which often arise when retrieval matches plausible motion to an incompatible 3D base, as seen in the second input of the third row, where a seven-drawer table’s motion is consistent but the geometry fails to accommodate it. \textbf{The last row} shows zero-shot results on in-the-wild or internet images, confirming our method’s capability on unconstrained inputs, underscoring both its strengths and current boundaries.

\end{document}